\documentclass[10pt,twocolumn,letterpaper]{article}

\usepackage{cvpr}              

\usepackage[dvipsnames]{xcolor}

\usepackage{amsmath}
\usepackage{amssymb}
\usepackage{booktabs}
\usepackage{physics}
\usepackage{array} 

\usepackage{algorithm}
\usepackage{algcompatible}
\usepackage{cite}
\usepackage{bbm}
\usepackage{multirow}
\usepackage{colortbl}
\usepackage{xcolor}
\usepackage{color}
\usepackage{adjustbox}
\usepackage{physics}
\usepackage{svg}
\usepackage{amsmath,bm}
\usepackage{dirtytalk}
\usepackage{lipsum}
\usepackage{setspace}

\DeclareMathOperator*{\softmax}{\mathtt{softmax}}
\DeclareMathOperator*{\argmax}{\mathtt{argmax}}

\definecolor{cvprblue}{rgb}{0.21,0.49,0.74}
\usepackage[pagebackref,breaklinks,colorlinks,citecolor=cvprblue]{hyperref}
\usepackage{makecell}
\usepackage{microtype}
\usepackage[accsupp]{axessibility} 
\usepackage{xcolor}

\usepackage[para, symbol*]{footmisc}

\usepackage[resetlabels]{multibib}

\newcites{Supp}{References}
\def\paperID{2273} 
\def\confName{CVPR}
\def\confYear{2024}

\title{Learning CNN on ViT: A Hybrid Model to Explicitly Class-specific Boundaries for Domain Adaptation}

\author{Ba Hung Ngo$^{1, }$\thanks{Co-first author.}  , Nhat-Tuong Do-Tran$^{2, \textcolor{red}{*}}$, Tuan-Ngoc Nguyen$^{3}$, Hae-Gon Jeon$^{4}$, Tae Jong Choi$^{1, }$\thanks{Corresponding author.}\\
$^{1}$Graduate School of Data Science, Chonnam National University, South Korea\\
$^{2}$Department of Computer Science, National Yang Ming Chiao Tung University, Taiwan\\
$^{3}$Digital Transformation Center, FPT Telecom, VietNam, 
$^{4}$AI Graduate School, GIST, South Korea\\
{\tt\footnotesize $\text{ngohung}$@chonnam.ac.kr \; $\text{tuongdotn.cs11}$@nycu.edu.tw \; $\text{tuannn55}$@fpt.com \; $\text{haegonj}$@gist.ac.kr \; $\text{ctj17}$@jnu.ac.kr}
}

\begin{document}
\maketitle
\begin{abstract} 
Most domain adaptation (DA) methods are based on either a convolutional neural networks (CNNs) or a vision transformers (ViTs). They align the distribution differences between domains as encoders without considering their unique characteristics. For instance, ViT excels in accuracy due to its superior ability to capture global representations, while CNN has an advantage in capturing local representations. 
This fact has led us to design a hybrid method to fully take advantage of both ViT and CNN, called \textbf{E}xplicitly \textbf{C}lass-specific \textbf{B}oundaries (\textbf{ECB}). 
ECB learns CNN on ViT to combine their distinct strengths. In particular, we leverage ViT's properties to explicitly find class-specific decision boundaries by maximizing the discrepancy between the outputs of the two classifiers to detect target samples far from the source support. 
In contrast, the CNN encoder clusters target features based on the previously defined class-specific boundaries by minimizing the discrepancy between the probabilities of the two classifiers. Finally, ViT and CNN mutually exchange knowledge to improve the quality of pseudo labels and reduce the knowledge discrepancies of these models.
Compared to conventional DA methods, our ECB achieves superior performance, which verifies its effectiveness in this hybrid model. The project website can be found \url{https://dotrannhattuong.github.io/ECB/website}. \vspace{-1em}

\end{abstract}    
\section{Introduction} \label{sec:introduction}

\begin{figure}[!ht]
\centering
\includegraphics[width=225pt]{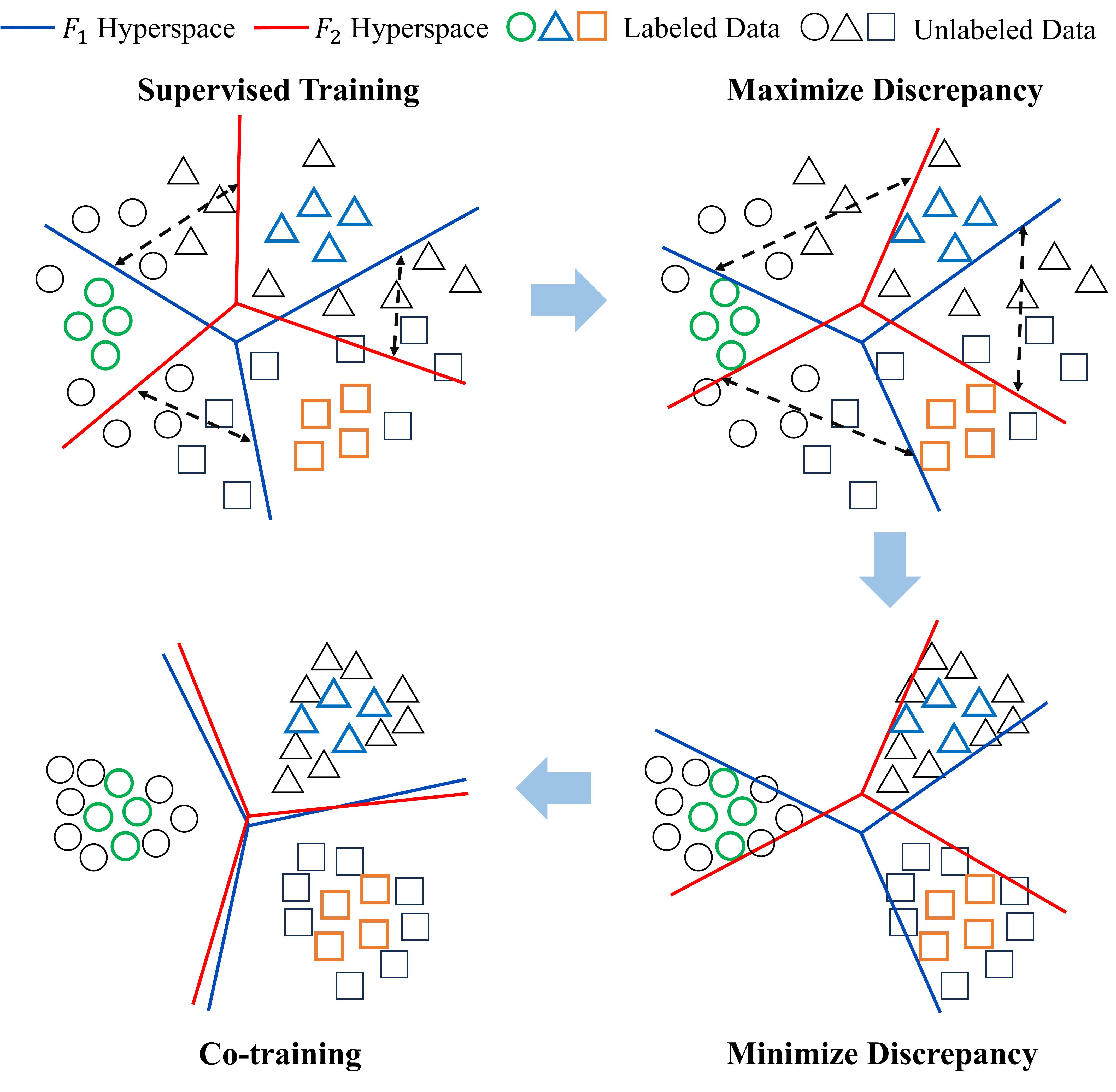}
\caption{Illustration of the proposed hybrid network architecture that leverages the strengths of ViT and CNN models.}  \label{fig_1} \vspace{-1.5em}
\end{figure}

Over the past few years, convolutional neural networks (CNNs) \cite{CNN} have been the cornerstone of deep learning techniques in computer vision tasks. This progress is mainly attributed to a convolution layer, which efficiently captures local spatial hierarchies for robust image representations. This local feature extraction capability has enabled CNNs to achieve State-of-the-Art (SOTA) performance in a variety of vision tasks, from image classification to object detection. In spite of its powerful local feature extraction, CNNs are somewhat limited in capturing more global and comprehensive visual context. To overcome this limitation, vision transformer (ViT) \cite{ViT} was introduced. ViT starts by dividing an image into patches, which are transformed into a sequence of tokens. Positional embeddings are incorporated into the tokens to retain the order of these patches. The model then uses transformer blocks to extract these tokens into features as image representations. Thanks to self-attention mechanisms in ViT, it is able to weigh the importance of different regions in an image irrespective of their spatial proximity, leading to comprehensive global representations with impressive accuracy. The advancements in ViT have led to a growing inclination in the machine learning realm over CNN-based approaches for a range of tasks. 

Instead of focusing solely on replacing CNN with ViT, our approach diverges from this trend. We believe that both ViT and CNN architectures have their own strengths that can be harnessed when combined well. For instance, ViT has shown the capability to capture global representations and demonstrates robustness when trained on large datasets. Yet, because it is composed of multilayer perceptron (MLP) layers, ViT can overfit if the dataset is limited. On the other hand, CNNs perform well with relatively small datasets thanks to their spatial invariance and robustness in capturing local representations. Motivated by this, we propose a novel method that capitalizes on the distinct strengths of both architectures. As shown in \cref{fig_1}, we exploit the superior accuracy of ViT in identifying more general class-specific boundaries by maximizing the discrepancy between the outputs of the two classifiers, enabling us to estimate the worst-case hyperspaces. Once these class-specific boundaries are defined, CNN minimizes the discrepancy by clustering target features closer to the source domain, aiming to minimize errors within the identified hyperspaces. However, the knowledge discrepancies between ViT and CNN still exist, so we applied additional co-training to bridge this gap while improving the quality of the pseudo labels. In the field of unsupervised domain adaptation (UDA), MCD \cite{MCD} emphasizes the importance of maximizing the discrepancy between classifier outputs for target samples, which are far from the source domain's support, and then minimizing this discrepancy through a feature generator to align the target features closer to the source's support. 

Although UDA has made significant advancements in domain adaptation (DA) tasks, the semi-supervised domain adaptation (SSDA) scenarios, as discussed in \cite{MME, APE, DECOTA, CDAC, S3D, ASDA, SLA, ProML, DARK}, are extensively employed to yield remarkable classification accuracy compared to the UDA setting \cite{MDD, GVB, BNM, SCDA, DALN}. This is because a model trained under UDA is only accessed to labeled source data, while a model trained with the SSDA setting benefits from the extra target information with a few labeled target samples besides labeled source data. However, the previous DA methods only take full advantage of the unique benefits of CNN as a feature extractor. Specifically, works in \cite{DAN, MCC, MME, APE} use a combination of a CNN encoder followed by an MLP classifier, but the decision boundary towards the source domain leads to data bias in DA. To address this bias, some previous works \cite{MDD, UODA, ASDA} introduce a multi-model by adding one more MLP classifier, which consists of a single CNN encoder and two MLP classifiers. Furthermore, the first DA approaches in \cite{FixBi, DECOTA} use two CNN encoders and two MLP classifiers to boost the classification accuracy by leveraging a co-training strategy that ensures the consistency of unlabeled target data through knowledge exchange. However, these methodologies still follow the same rule, where the CNN model is selected as the encoder, and MLP is assigned as the classifier. Therefore, the capabilities of ViT in capturing global information remain unexploited. Inspired by the ideal, we make use of this multi-model architecture to build a new hybrid framework that leverages the advantages of ViT and CNN for DA settings. However, the proposed method does not require any additional complexity compared to previous works in the test phase. We summarize the contributions of this paper as follows: 
\begin{itemize}
    \item We introduce a hybrid model that can take advantage of ViT. Beyond simply replacing CNN with ViT, we can drive the feature space of ViT to CNN. 
    
    \item Our approach demonstrates the successful integration of ViT and CNN, making a synergy with these two powerful frameworks. 
    
    \item The proposed method outperforms the prior works to achieve SOTA performances on DA benchmark datasets. 
\end{itemize}
\section{Related Works} \label{sec:relatedworks}

\textbf{Co-training.} In the realm of semi-supervised learning (SSL), co-training is a scheme to improve the robustness, first proposed by \cite{blum1998combining}, and harnesses data from dual views, enabling two models to iteratively 'teach' each other. During this process, each model alternately makes predictions on unlabeled data, with the most confident predictions used to augment the training set of the other model. This mutual teaching strategy can significantly improve the performance of both models, particularly when labeled data is scarce. In the DA context, FixBi \cite{FixBi}, DECOTA \cite{DECOTA}, and MVCL \cite{MVCL} offer innovative co-training strategies. Notably, FixBi and DECOTA utilize two distinct branches, each including a feature extractor and a classifier. They utilize MIXUP augmentation to reduce the gap in multiple intermediate domains between the source and target domain during co-training. However, they miss the potential benefits of strong augmentation. In contrast, while MVCL employs both weak and strong augmentation to bolster its co-training approach, it relies on a single CNN-based encoder and two classifiers to produce two views. This limits the representation of unlabeled target data, depriving its comprehensive global information. To address these limitations, we employ the ViT in conjunction with a CNN-based encoder. This combination generates two representational views that encapsulate both local and global information. Additionally, we integrate both weak and strong augmentations for unlabeled target samples, which enhances the interactivity and effectiveness of our co-training strategy.

\begin{figure*}[!t]
 \begin{center}
    \includegraphics[width=495pt]{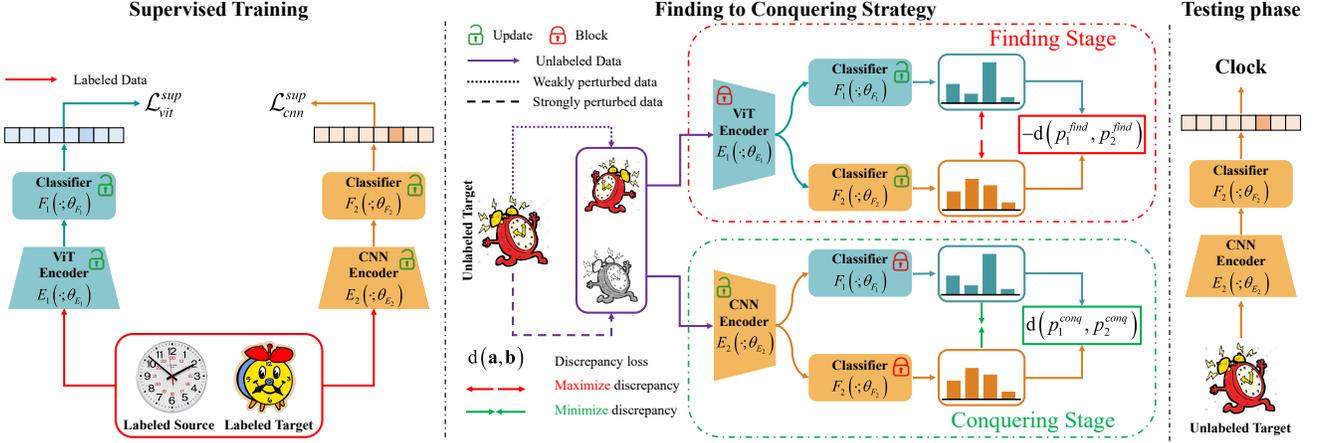}
    \caption{Illustration of a hybrid network with the proposed Finding to Conquering strategy. We use ViT to build $E_{1}$ that drives two classifiers $F_{1}$ and $F_{2}$ to expand class-specific boundaries comprehensively. Besides, we select CNN for the second encoder $E_{2}$ to cluster target features based on the boundaries identified by ViT. These encoders all use two classifiers $F_1$, $F_2$.}  \label{fig_2} \vspace{-2.0em}
 \end{center}
\end{figure*}

\noindent\textbf{Domain Adaptation Framework.} In the realm of DA, various frameworks have been presented. Early methods such as MME \cite{MME}, APE \cite{APE}, and SENTRY \cite{SENTRY} adopt the conventional approach of constructing deep learning frameworks that have a feature extractor coupled with a classifier. However, source and target domains share the same decision boundary, which leads to data bias toward the source domain. To alleviate data bias, MDD \cite{MDD} and UODA \cite{UODA} introduce two distinct classifiers. They demonstrate that using a dual-classifier setup can improve the performance of classification tasks. These methods use a single feature extractor to train on both labeled source and target datasets. However, the source domain is more reliable than the target domain thanks to labeled source samples, which can lead the feature extractor to overly focus on source data representations. This results in a learning bias toward the source domain, accumulating errors during the training process. To solve this problem, DECOTA \cite{DECOTA} proposes a novel architecture by decomposing two distinct branches, UDA and SSL, each consisting of a feature extractor and a single classifier. Specifically, the UDA branch is trained on labeled source data and unlabeled target data, while the SSL branch is trained on labeled and unlabeled target data. Thanks to the SSL branch, focusing solely on the target domain facilitates the alleviation of learning bias. In addition, the UDA branch is capable of mitigating learning bias by leveraging extra information from the SSL branch via the co-training strategy. Despite the notable advancements in the field, DECOTA still follows the framework of using CNN as the feature extractor. Therefore, it is impossible to completely improve the quality of pseudo labels for unlabeled target samples and exploit learning space based on the properties of ViT. As a result, the capabilities of ViT in capturing global information still need to be explored. Instead of solely replacing CNN with ViT, we propose a hybrid model that combines their strengths of both architectures. 
\section{Methodology} \label{sec:method}
In DA scenarios, we deal with the following data setting:
\begin{itemize}
  \item \textbf{Labeled source domain}: Denoted as $\mathcal{D_S} = \{(x^{\mathcal{S}}_i, y^{\mathcal{S}}_i)\}^{\mathcal{N_S}}_{i=1}$ includes $\mathcal{N_S}$ richly labeled samples.
  \item \textbf{Labeled target domain}: Denoted as $\mathcal{D}_{\mathcal{T}_l} = \{{(x^{\mathcal{T}_l}_i, y^{\mathcal{T}_l}_i)\}}^{\mathcal{N}_{\mathcal{T}_l}}_{i=1}$ includes $\mathcal{N}_{\mathcal{T}_l}$ limited labeled target samples. \textit{Notably, $\mathcal{D}_{\mathcal{T}_l}$ is empty in UDA}.
  \item \textbf{Unlabeled target domain}: Denoted as $\mathcal{D}_{\mathcal{T}_u} = \{{(x^{\mathcal{T}_u}_i, y^{\mathcal{T}_u}_i)\}}^{\mathcal{N}_{\mathcal{T}_u}}_{i=1}$ includes $\mathcal{N}_{\mathcal{T}_u}$ target samples that do not have corresponding labels during the training phase.
\end{itemize}

In this setup, the sample $x^{\mathcal{S}}_i$ and $x^{\mathcal{T}_l}_i$ from the source domain and the labeled target domain are associated with corresponding ground-truth labels $y^{\mathcal{S}}_i$ and $y^{\mathcal{T}_l}_i$, respectively. There is an assumption that the label $y^{\mathcal{S}}$, $y^{\mathcal{T}_l}$, and $y^{\mathcal{T}_u}$ all belong to the same label space with $K$ classes. Notably, $y^{\mathcal{T}_u}$, which denotes labels for the unlabeled target data, are only used during the testing phase. Furthermore, the number of labeled target samples $\mathcal{N}_{\mathcal{T}_l}$ is much smaller than both $\mathcal{N_S}$ and $\mathcal{N}_{\mathcal{T}_u}$. Moreover, we implement two varied stochastic data transformations: $Aug_w(\cdot)$ and $Aug_{str}(\cdot)$. The function $Aug_w(\cdot)$ is a weak augmentation method employing light perturbations such as random horizontal flipping and random cropping, whereas $Aug_{str}(\cdot)$ stands as a strong augmentation method, using RandAugment \cite{Randaugment}, which involves 14 transformation techniques. Specifically, both $Aug_w(\cdot)$ and $Aug_{str}(\cdot)$ are applied to the unlabeled set $\mathcal{D}_{\mathcal{T}_u}$, transforming a sample $x^{\mathcal{T}_u}_{i}$ to two versions $x^{\mathcal{T}_u}_{i,w}$ and $x^{\mathcal{T}_u}_{i,{str}}$, respectively. Subsequently, we train the model on the labeled set $\mathcal{D}_l=\mathcal{D_S} \cup \mathcal{D}_{\mathcal{T}_l}$ and the unlabeled set $\mathcal{D}_{\mathcal{T}_u}$, and evaluate the trained model on $\mathcal{D}_{\mathcal{T}_u}$.

To improve the performance on $\mathcal{D}_{\mathcal{T}_u}$, we propose a hybrid model consisting of ViT and CNN branches. The ViT branch is made up of a ViT encoder ${E_1}( \cdot ;{\boldsymbol{\theta}_{{E_1}}})$ and a classifier ${F_1}( \cdot ;{\boldsymbol{\theta}_{{F_1}}})$, while the CNN branch includes a CNN encoder ${E_2}( \cdot ;{\boldsymbol{\theta}_{{E_2}}})$ and a classifier ${F_2}( \cdot ;{\boldsymbol{\theta}_{{F_2}}})$. Our strategy training proceeds in three steps:
\begin{enumerate}
    \item \textbf{Supervised Training}: We train both ViT and CNN branches on $\mathcal{D}_l$ whose knowledge is adapted to the $\mathcal{D}_{\mathcal{T}_u}$ as illustrated in \cref{fig_2}.
    \item \textbf{Finding to Conquering}: In \cref{fig_2}, we find class-specific boundaries based on fixed $E_{1}$ by maximizing discrepancy between $F_{1}$ and $F_{2}$. Subsequently, the $E_{2}$ clusters the target features based on those class-specific boundaries by minimizing discrepancy.
    \item \textbf{Co-training}: To exchange effectively knowledge between two branches on $\mathcal{D}_{\mathcal{T}_u}$, the ViT branch generates a pseudo label for weakly version $x^{\mathcal{T}_u}_{i,w}$ to teach the CNN branch with strongly version $x^{\mathcal{T}_u}_{i, str}$. Then, the reverse process is also applied as depicted in \cref{fig_3}.
\end{enumerate}

\subsection{Supervised Training} 
In this phase, we employ the standard cross-entropy loss to train a model's two branches: \textbf{ViT branch} and \textbf{CNN branch}. In the ViT branch, we aim to minimize the empirical loss of labeled data, which is defined as follows: 
\begin{equation} \label{eqn_1}
    \mathcal{L}^{sup}_{vit}({\boldsymbol{\theta}_{{E_1}}}, {\boldsymbol{\theta}_{{F_1}}}) = \frac{1}{\mathcal{N}_l}\sum^{\mathcal{N}_l}_{i=1}H(y^l_{i}, p^l_1(x^l_i)), 
\end{equation} 
\noindent where $H(\cdot)$ denotes the standard cross-entropy loss. When $p^l_1(x^l_i) = \sigma({F_1}(E_1(x^l_i)))$, $\sigma$ represents the $\softmax$ function, which transforms the logits into probabilities across $K$ categories. Initially, ViT encoder $E_1$ maps the labeled sample $x^l_i$ into a $d$-dimensional feature space. The classifier $F_1$ then categorizes the input feature space $E_1(x^l_{i})$ into the logits.  Finally, we minimize the cross-entropy loss between the predicted labeled probability by $p^l_1(x^l_i)$ and the provided ground-truth label $y^l_{i}$, ensuring accurate predictions on labeled data.

Similarly, the CNN branch is also trained using labeled data to minimize the empirical loss as follows: 
\begin{equation} \label{eqn_2}
    \mathcal{L}^{sup}_{cnn}({\boldsymbol{\theta}_{{E_2}}}, {\boldsymbol{\theta}_{{F_2}}}) = \frac{1}{\mathcal{N}_l}\sum^{\mathcal{N}_l}_{i=1}H(y^l_{i}, p^l_2(x^l_i)), 
\end{equation}
\noindent where $p^l_2(x^l_i)=\sigma({F_2}(E_2(x^l_i)))$ is obtained by extracting features from the labeled sample $x^l_i$ using the CNN encoder $E_2$. These features are then mapped to logits through the classifier $F_2$. Finally, the $\softmax$ function $\sigma$ transforms these logits into probabilities across $K$ categories. The main goal of this process is to minimize the cross-entropy loss to ensure a close alignment between the predicted probabilities $p^l_2(x^l_i)$ and the ground-truth labels $y^l_{i}$.

\subsection{Finding to Conquering} 
\textbf{Discrepancy Loss Definition.} In this work, we use the absolute difference between the probabilistic outputs $\textbf{a}$ and $\textbf{b}$ from two distinct classifiers $F_1$ and $F_2$, respectively, for each class to guarantee that the discrepancy loss is always non-negative as follows:
\begin{equation} \label{eqn_3} 
    d(\textbf{a},\textbf{b}) = \frac{1}{K}\sum^{K}_{k=1}|a_{k} - b_{k}|, 
\end{equation}
\noindent where $a_{k}$ and $b_{k}$ correspond to the probability outputs of the two classifiers for the $k$-th class. Normalizing with the total number of categories assures that the discrepancy loss is scale-invariant and bounded between 0 and 1. The discrepancy loss serves as a measure of divergence between the probabilistic outputs of two classifiers, $\textbf{a}$ and $\textbf{b}$.  A discrepancy loss of zero indicates the perfect agreement between them across all classes. On the other hand, a higher discrepancy loss points to a more significant divergence in the two classifiers's predictions. This loss is beneficial in scenarios where we want to expand class-specific boundaries.

\noindent\textbf{Finding Stage.} In this stage, we aim to expand the class-specific boundaries of the classifiers $(F_1, F_2)$ in relation to the fixed ViT encoder $E_1$ by maximizing the discrepancy loss between their outputs. This means we have a possibility to estimate worst-case hyperspaces, which identifies target samples that fall outside the broader source distribution's support, rather than using CNN encoder $E_2$. To avoid this issue, we add a supervised loss on the labeled samples as follows:
\begin{center}
$\mathcal{L}_{find}({\boldsymbol{\theta}_{{F_1}}}, {\boldsymbol{\theta}_{{F_2}}})
= \mathcal{L}^{sup}_{vit} + \mathcal{L}^{sup}_{cnn}$ 
\end{center}
\begin{equation} \label{eqn_4}
\begin{gathered}
- \frac{1}{\mathcal{N}_{\mathcal{T}_u}}\sum^{\mathcal{N}_{\mathcal{T}_u}}_{i=1} d\big(p^{find}_1(x^{\mathcal{T}_u}_{i}), p^{find}_2(x^{\mathcal{T}_u}_{i})\big),
\end{gathered}
\end{equation}
\noindent where $p^{find}_1(x^{\mathcal{T}_u}_{i}) = \sigma(F_{1}(E_{1}(x^{\mathcal{T}_u}_{i})))$ and $p^{find}_2(x^{\mathcal{T}_u}_{i}) = \sigma(F_{2}(E_{1}(x^{\mathcal{T}_u}_{i})))$ denote the probability outputs of $F_{1}$ and $F_{2}$ with ViT encoder $E_1$ on the unlabeled target data $x^{\mathcal{T}_u}_{i}$, respectively. By integrating the supervised loss on the labeled samples, we ensure that the classifiers not only distinguish different classes, but also push to deviate as far as possible from the currently learned class-specific boundaries while all labeled samples are correctly distinguished. In essence, the Finding Stage is a delicate balance between maximizing the discrepancy to explore the class-specific boundaries and minimizing the supervised loss to maintain classification accuracy.

\noindent\textbf{Conquering Stage.} In this stage, we leverage the class-specific boundaries established by the ViT encoder $E_1$ as a reference to guide the optimization of the CNN encoder $E_2$, while keeping the classifiers $F_1$ and $F_2$ fixed. The primary is to minimize the discrepancy between the outputs of two classifiers, $F_1$ and $F_2$, allowing the features extracted by CNN encoder $E_2$ to be accurately classified. The objective function is formulated as follows:
\begin{equation} \label{eqn_5}
    \mathcal{L}_{conq}({\boldsymbol{\theta}_{{E_2}}}) = \frac{1}{\mathcal{N}_{\mathcal{T}_u}}\sum^{\mathcal{N}_{\mathcal{T}_u}}_{i=1} d\big(p^{conq}_1(x^{\mathcal{T}_u}_{i}), p^{conq}_2(x^{\mathcal{T}_u}_{i})\big),
\end{equation}
\noindent where $p^{conq}_1(x^{\mathcal{T}_u}_{i}) = \sigma(F_{1}(E_{2}(x^{\mathcal{T}_u}_{i})))$ and $p^{conq}_2(x^{\mathcal{T}_u}_{i}) = \sigma(F_{2}(E_{2}(x^{\mathcal{T}_u}_{i})))$ denote the probability outputs of $F_{1}$ and $F_{2}$ with CNN encoder $E_2$ on the unlabeled target data, $x^{\mathcal{T}_u}_{i}$. By focusing on minimizing the discrepancy loss, the features extracted by the CNN encoder $E_2$ are clustered on the class-specific boundaries of the ViT encoder $E_1$, even when estimating the worst-case hyperspaces. This alignment is crucial as it guarantees that the unlabeled samples are not only correctly classified but also robust to variations in the class-specific boundaries.

\begin{figure}[!t]
 \begin{center}
    \includegraphics[width=230pt]{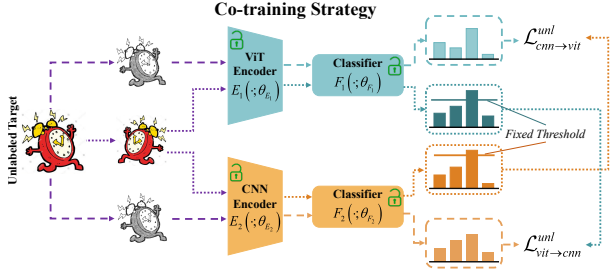}
    \caption{Illustration of co-training strategy.}  \label{fig_3} \vspace{-2.0em}
 \end{center}
\end{figure}

\subsection{Co-training} 
Following the Finding-to-Conquering (FTC) strategy, we recognize a significant gap in the knowledge discrepancy between the ViT and CNN branches that needs to be minimized for the optimal performance. To resolve this, we have adopted a co-training strategy with dual objectives as illustrated in \cref{fig_3}. The first objective focuses on reducing the gap between the two branches by enabling mutual enhancement and leveraging each branch's strength to improve the quality of pseudo labels. The second objective is to specifically improve the performance of the CNN branch, thanks to the potential to capture complex patterns and relationships in the data of the ViT branch.

Initially, we employ pseudo labels generated by the ViT branch to teach the learning process of the CNN branch. This is achieved by setting a fixed threshold, denoted as $\tau_{vit}$, and applying it as follows: 
\begin{center}
$\mathcal{L}^{unl}_{{vit}\rightarrow{cnn}}({\boldsymbol{\theta}_{{E_2}}}, {\boldsymbol{\theta}_{{F_2}}})
=$ 
\end{center}
\begin{equation} \label{eqn_6}
\begin{gathered}
\frac{1}{\mathcal{N}_{\mathcal{T}_u}}\sum^{\mathcal{N}_{\mathcal{T}_u}}_{i=1} \mathbbm{1}[\max(\textbf{q}^v_i) \geq \tau_{vit}] H\big(\hat{q}^{v}_{i}, p^c(x^{\mathcal{T}_u}_{i, str})\big),
\end{gathered}
\end{equation}
\noindent where $\mathbbm{1}[\cdot]$ is the binary indicator, returning 1 if the condition inside [$\cdot$] is satisfied, and 0 otherwise. In this context, $\textbf{q}^{v}_{i} = \sigma({F_1}(E_{1}(x^{\mathcal{T}_u}_{i,w})))$ indicates the ViT branch's prediction for a weakly augmented version of an unlabeled target sample. Then, the highest prediction, exceeding the fixed threshold $\tau_{vit}$, is converted into pseudo label $\hat{q}^{v}_{i} = \argmax(\textbf{q}^{v}_{i})$. Finally, the output prediction $p^c(x^{\mathcal{T}_u}_{i, str}) = \sigma({F_2}(E_{2}(x^{\mathcal{T}_u}_{i, str})))$ of the CNN branch on the strongly augmented target sample is adjusted to closely align with the pseudo label $\hat{q}^{v}_{i}$ using the cross-entropy loss. This process ensures that the CNN branch effectively learns from the ViT branch's predictions, making the consistency and alignment between the two branches better.

Similarly, we also leverage the pseudo labels from the CNN branch to guide the learning process of the ViT branch. This is achieved by setting another fixed threshold, denoted as $\tau_{cnn}$, and using it as follows: 
\begin{center}
$\mathcal{L}^{unl}_{{cnn}\rightarrow{vit}}({\boldsymbol{\theta}_{{E_1}}}, {\boldsymbol{\theta}_{{F_1}}})
=$
\end{center}
\begin{equation} \label{eqn_7}
\begin{gathered}
\frac{1}{\mathcal{N}_{\mathcal{T}_u}}\sum^{\mathcal{N}_{\mathcal{T}_u}}_{i=1}\mathbbm{1}[\max(\textbf{q}^c_i) \geq \tau_{cnn}]H\big(\hat{q}^{c}_{i}, p^v(x^{\mathcal{T}_u}_{i, str})\big),
\end{gathered}
\end{equation}
\noindent where $\textbf{q}^{c}_{i} = \sigma(F_2(E_{2}(x^{\mathcal{T}_u}_{i,w})))$ is the CNN branch's prediction for a weakly augmented target sample. The pseudo label $\hat{q}^{c}_{i}$ is generated from the highest predictions of the CNN branch that is higher than the threshold $\tau_{cnn}$. Meanwhile, $p^v(x^{\mathcal{T}_u}_{i, str}) = \sigma(F_1(E_{1}(x^{\mathcal{T}_u}_{i, str})))$ indicates the output prediction of the ViT branch on the strongly augmented target sample and is adjusted to closely align with the pseudo label $\hat{q}^{c}_{i}$, using the cross-entropy loss. Through the co-training process, we enhance the model's ability to generalize and perform accurately on unlabeled target data.

\subsection{Testing Phase} 
For a fair comparison with the previous DA methods \cite{MCC, SENTRY, MME, CDAC, SLA}, we select the CNN encoder $E_{2}$ associated with its classifier $F_{2}$ in the testing phase as illustrated in \cref{fig_2} as follows:
\begin{equation}\label{E11}
    \hat{y}^{\mathcal{T}_u}_{i} = \argmax \big(F_{2}(E_{2}(x^{\mathcal{T}_u}_{i}))\big), 
\end{equation}
\noindent where $\hat{y}^{\mathcal{T}_u}_{i}$ is the predicted class of unlabeled target sample.
\section{Experiments} \label{sec:experiments}
\subsection{Experiment Setup}
\textbf{Datasets}. We conduct extensive evaluations on standard DA benchmark datasets: \textit{Office-Home} \cite{Office-Home} and \textit{DomainNet} \cite{DomainNet}. On \textit{Office-Home} dataset, we perform experiments on all possible combinations of these 4 domains: Real ($R$), Clipart ($C$), Art ($A$), and Product ($P$) with 65 categories. Consistent with prior SSDA methods \cite{MME, CDAC, SLA, ProML, G-ABC}, we conduct an evaluation on a subset of \textit{DomainNet} that includes 126 categories in 4 domains: Real ($rel$), Clipart ($clp$), Painting ($pnt$), and Sketch ($skt$) using 7 different mixtures of these domains. 

\begin{table*}[ht]
\resizebox{\textwidth}{!}{%
\begin{tabular}{c|cccccccccccc|c} \toprule
Method & $A{\rightarrow}C$ & $A{\rightarrow}P$ & $A{\rightarrow}R$ & $C{\rightarrow}A$ & $C{\rightarrow}P$ & $C{\rightarrow}R$ & $P{\rightarrow}A$ & $P{\rightarrow}C$ & $P{\rightarrow}R$ & $R{\rightarrow}A$ & $R{\rightarrow}C$ & $R{\rightarrow}P$ & Mean \\ \midrule
DANN \cite{DANN}      &     45.6      &     59.3      &     70.1      &     47.0      &     58.5      &     60.9      &     46.1      &     43.7      &     68.5      &     63.2      &     51.8      &     76.8      &     57.6       \\

MCD \cite{MCD}        &     48.9      &     68.3      &     74.6      &     61.3      &     67.6      &     68.8      &     57.0      &     47.1      &     75.1      &     69.1      &     52.2      &     79.6      &     64.1       \\

BNM \cite{BNM}        &     52.3      &     73.9      &     80.0      &     63.3      &     72.9      &     74.9      &     61.7      &     49.5      &     79.7      &     70.5      &     53.6      &     82.2      &     67.9        \\

MDD \cite{MDD}        &     54.9      &     73.7      &     77.8      &     60.0      &     71.4      &     71.8      &     61.2      &     53.6      &     78.1      &     72.5      &     60.2      &     82.3      &     68.1       \\

MCC \cite{MCC}        &     55.1      &     75.2      &     79.5      &     63.3      &     73.2      &     75.8      &     66.1      &     52.1      &     76.9      &     73.8      &     58.4      &     83.6      &     69.4       \\

GVB \cite{GVB}        &     57.0      &     74.7      &     79.8      &     64.6      &     74.1      &     74.6      &     65.2      &     55.1      &     81.0      &     74.6      &     59.7      &     84.3      &     70.4       \\

DCAN \cite{DCAN}      &     54.5      &     75.7      &     81.2      &     67.4      &     74.0      &     76.3      &     67.4      &     52.7      &     80.6      &     74.1      &     59.1      &     83.5      &     70.5       \\

DALN \cite{DALN}      &     57.8      &     79.9      &     82.0      &     66.3      &     76.2      &     77.2      &     66.7      &     55.5      &     81.3      &     73.5      &     60.4      &     85.3      &     71.8       \\


FixBi \cite{FixBi}    &     58.1      &     77.3      &     80.4      &     67.7      &     79.5      &     78.1      &     65.8      &     57.9      &     81.7      &     76.4      &     62.9      &     86.7      &     72.7       \\

DCAN+SCDA \cite{SCDA} &     60.7      &     76.4      &     \underline{82.8}        &     69.8      &     77.5      &     78.4      &     68.9      &     59.0      &     82.7      &     74.9      &     61.8      &     84.5      &     73.1       \\

ATDOC \cite{ATDOC}    &     60.2      &     77.8      &     82.2      &     68.5      &     78.6      &     77.9      &     68.4      &     58.4      &     83.1      &     74.8      &     61.5      &     \underline{87.2}      &     73.2       \\ 

EIDCo \cite{EIDCo}    &     \underline{63.8}      &     \underline{80.8}      &     82.6      &     \underline{71.5}      &     \underline{80.1}      &     \underline{80.9}      &     \underline{72.1}      &     \underline{61.3}      &     \underline{84.5}      &     \underline{78.6}      &     \underline{65.8}      &     87.1      &     \underline{75.8}       \\

\textbf{ECB (CNN)}  & \textbf{68.5} & \textbf{85.4} & \textbf{88.3} & \textbf{79.2} & \textbf{86.8} & \textbf{89.0} & \textbf{79.3} & \textbf{66.4} & \textbf{88.5} & \textbf{81.0} & \textbf{71.1} & \textbf{90.4} & \textbf{81.2}   \\  \bottomrule
\end{tabular} 
} \vspace{-0.5em}
\caption{\textbf{Accuracy (\%) on Office-Home} of UDA methods across different domain shifts. \textbf{ECB (CNN)} represents the performance of our method when applied to ResNet-50. The top and second-best accuracy results are highlighted in \textbf{bold} and \underline{underline} for easy identification.} \vspace{-0.25em}\label{result_officehome_uda} 
\end{table*}

\begin{table*}[!t]
\resizebox{\textwidth}{!}{
\begin{tabular}{c|cccccccccccccc|cc}
\toprule
&
\multicolumn{2}{c}{$rel{\rightarrow}clp$} &
\multicolumn{2}{c}{$rel{\rightarrow}pnt$} &
\multicolumn{2}{c}{$pnt{\rightarrow}clp$} &
\multicolumn{2}{c}{$clp{\rightarrow}skt$} &
\multicolumn{2}{c}{$skt{\rightarrow}pnt$} &
\multicolumn{2}{c}{$rel{\rightarrow}skt$} &
\multicolumn{2}{c|}{$pnt{\rightarrow}rel$} &
\multicolumn{2}{c}{Mean} \\ \vspace{0.1em}
\multirow{-2}{*}{Method} &
\multicolumn{1}{l}{1\textsubscript{shot}} &
\multicolumn{1}{l}{3\textsubscript{shot}} &
\multicolumn{1}{l}{1\textsubscript{shot}} &
\multicolumn{1}{l}{3\textsubscript{shot}} &
\multicolumn{1}{l}{1\textsubscript{shot}} &
\multicolumn{1}{l}{3\textsubscript{shot}} &
\multicolumn{1}{l}{1\textsubscript{shot}} &
\multicolumn{1}{l}{3\textsubscript{shot}} &
\multicolumn{1}{l}{1\textsubscript{shot}} &
\multicolumn{1}{l}{3\textsubscript{shot}} &
\multicolumn{1}{l}{1\textsubscript{shot}} &
\multicolumn{1}{l}{3\textsubscript{shot}} &
\multicolumn{1}{l}{1\textsubscript{shot}} &
\multicolumn{1}{l|}{3\textsubscript{shot}} &
\multicolumn{1}{l}{1\textsubscript{shot}} &
\multicolumn{1}{l}{3\textsubscript{shot}} \\ \midrule

ENT \cite{ENT}          &     65.2      &     71.0      &     65.9      &     69.2      &     65.4      &     71.1      &     54.6      &     60.0      &     59.7      &     62.1      &     52.1      &     61.1      &     75.0      &     78.6      &     62.6      &     67.6       \\

MME \cite{MME}          &     70.0      &     72.2      &     67.7      &     69.7      &     69.0      &     71.7      &     56.3      &     61.8      &     64.8      &     66.8      &     61.0      &     61.9      &     76.1      &     78.5      &     66.4      &     68.9      \\

S$^3$D \cite{S3D}       &     73.3      &     75.9      &     68.9      &     72.1      &     73.4      &     75.1      &     60.8      &     64.4      &     68.2      &     70.0      &     65.1      &     66.7      &     79.5      &     80.3      &     69.9      &     72.1      \\

ATDOC \cite{ATDOC}      &     74.9      &     76.9      &     71.3      &     72.5      &     72.8      &     74.2      &     65.6      &     66.7      &     68.7      &     70.8      &     65.2      &     64.6      &     81.2      &     81.2      &     71.4      &     72.4       \\

MAP-F \cite{MAP-F}    &     75.3      &     77.0      &      74.0      &     75.0      &      74.3     &      77.0     &       65.8     &       69.5     &      73.0     &      73.3     &      67.5     &      69.2     &      81.7     &      83.3     &      73.1     &      74.9     \\

DECOTA \cite{DECOTA}    &     79.1      &     80.4      &     74.9      &     75.2      &      76.9     &      78.7     &      65.1     &      68.6     &      72.0     &      72.7     &      69.7     &      71.9     &      79.6     &      81.5     &      73.9     &      75.6     \\

CDAC \cite{CDAC}        &     77.4      &     79.6      &     74.2      &     75.1      &     75.5      &     79.3      &     67.6      &     69.9      &     71.0      &     73.4      &     69.2      &     72.5      &     80.4      &     81.9      &     73.6      &     76.0      \\

ASDA \cite{ASDA}        &     77.0      &     79.4      &     75.4      &     76.7      &      75.5     &      78.3     &      66.5     &      70.2     &      72.1     &      74.2     &      70.9     &      72.1     &      79.7     &      82.3     &      73.9     &      76.2     \\

CDAC+SLA \cite{SLA}     &     79.8      &     81.6      &     75.6      &     76.0      &      77.4     &      80.3     &      68.1     &      71.3      &      71.7     &      73.5     &      71.7     &      73.5     &      80.4     &      82.5     &      75.0     &      76.9     \\

ProML \cite{ProML}      &     78.5      &     80.2      &     75.4      &     76.5      &      77.8     &      78.9     &      70.2     &      72.0     &      74.1     &      75.4     &      72.4     &      73.5     &       \underline{84.0}     &      84.8     &      76.1     &      77.4     \\

MVCL \cite{MVCL}        &     78.8      &     79.8      &     76.0      &      \underline{77.4}      &      78.0     &     80.3     &      70.8     &      73.0     &       \underline{75.1}     &       \underline{76.7}     &      72.4     &       \underline{74.4}     &      82.4     &       \underline{85.1}     &      76.2     &      78.1     \\ 

G-ABC \cite{G-ABC}      &     \underline{80.7}      &      \underline{82.1}      &      \underline{76.8}      &     76.7      &      \underline{79.3}      &      \underline{81.6}      &      \underline{72.0}      &      \underline{73.7}      &     75.0      &     76.3      &      \underline{73.2}      &     74.3      &     83.4      &     83.9      &      \underline{77.2}      &      \underline{78.4}   \\

\textbf{ECB (CNN)}     & \textbf{83.8} & \textbf{87.4} & \textbf{85.4} & \textbf{85.6} & \textbf{86.4} & \textbf{87.3} & \textbf{79.7} & \textbf{80.6} & \textbf{83.4} & \textbf{85.6} & \textbf{79.5} & \textbf{81.7} & \textbf{88.7} & \textbf{90.3} & \textbf{83.8} & \textbf{85.5}  \\  \bottomrule
\end{tabular} 
} \vspace{-0.5em}
\caption{\textbf{Accuracy (\%) on DomainNet} of SSDA methods in both 1-shot and 3-shot settings using ResNet-34.}  \vspace{-1.5em}\label{result_domainnet}  
\end{table*}

\noindent\textbf{Implementation Details}. We use the ViT/B-16 \cite{ViT} for the ViT encoder $E_{1}$, while the ResNet \cite{ResNet-34} is adopted for the CNN encoder $E_{2}$. For UDA, we used ResNet-50 as the backbone network following previous works \cite{MCD, MDD, MCC, DALN}. Similarly, we follow the evaluation protocol of previous SSDA methods \cite{ENT, MME, ASDA, G-ABC}, ResNet-34 is applied for this scenario on the \textit{DomainNet} dataset. Each model is initially pre-trained on the ImageNet-1K \cite{ImageNet}. Given the disparate nature of the backbone architectures, we will have two different classifiers, $F_1$ and $F_2$, consisting of two fully-connected layers followed by the softmax function. We optimize our models using stochastic gradient descent with a $0.9$ momentum and $0.0005$ weight decay. Considering the different architectures of the ViT and the CNN branches, we set the initial learning rates for ViT and CNN at $1e-4$ and $1e-3$, respectively. The size of the mini-batch is set to $32$ for both $\mathcal{D}_l$ and $\mathcal{D}_{\mathcal{T}_u}$. The confidence threshold values for pseudo-label selection are set to $\tau_{vit}=0.6$ and $\tau_{cnn}=0.9$. The warmup phase for both branches on $\mathcal{D}_l$ is finetuned for $100,000$ iterations. Then, we update the learning rate scheduler and proceed with $50,000$ iterations of training for our method.

\begin{table*}[!t]
\resizebox{\textwidth}{!}{%
\centering
\begin{tabular}{ccccccccccccccccc} \toprule
\multicolumn{1}{c}{\multirow{2.5}{*}{Method}} & \multicolumn{2}{|c|}{$rel{\rightarrow}clp$} & \multicolumn{2}{c|}{$rel{\rightarrow}pnt$} & \multicolumn{2}{c|}{$pnt{\rightarrow}clp$} & \multicolumn{2}{c|}{$clp{\rightarrow}skt$} & \multicolumn{2}{c|}{$skt{\rightarrow}pnt$} & \multicolumn{2}{c|}{$rel{\rightarrow}skt$} & \multicolumn{2}{c|}{$pnt{\rightarrow}rel$}  & \multicolumn{2}{c}{Mean} \\ \cmidrule{2-17}

& \multicolumn{1}{|c}{ViT}       & \multicolumn{1}{c|}{CNN}           & ViT            & \multicolumn{1}{c|}{CNN}           & ViT            & \multicolumn{1}{c|}{CNN}           & ViT            & \multicolumn{1}{c|}{CNN}           & ViT            & \multicolumn{1}{c|}{CNN}           & ViT            & \multicolumn{1}{c|}{CNN}           & ViT            & \multicolumn{1}{c|}{CNN}           & ViT         & \multicolumn{1}{c}{CNN}        \\ \midrule 

\multicolumn{1}{c|}{\rule{0pt}{10pt}$vit{\rightarrow}cnn$}  &  73.3 & \multicolumn{1}{c|}{79.0} & 78.8 & \multicolumn{1}{c|}{81.0} & 75.1 & \multicolumn{1}{c|}{79.2} & 71.6 & \multicolumn{1}{c|}{74.7} & 78.6 & \multicolumn{1}{c|}{80.8} & 67.2 & \multicolumn{1}{c|}{72.0} & 88.1 & \multicolumn{1}{c|}{88.8} & 76.1 & 79.4 \\ 

\multicolumn{1}{c|}{\rule{0pt}{15pt}$cnn{\rightarrow}vit$}  & 74.2 & \multicolumn{1}{c|}{61.9} & 76.8 & \multicolumn{1}{c|}{66.8} & 76.1 & \multicolumn{1}{c|}{67.4} & 69.5 & \multicolumn{1}{c|}{57.2} & 74.9 & \multicolumn{1}{c|}{64.6} & 67.4 & \multicolumn{1}{c|}{54.8} & 86.0 & \multicolumn{1}{c|}{76.1} & 75.0 & 64.1 \\ \midrule

\multicolumn{1}{c|}{\rule{0pt}{15pt} co-training}  & 87.4 & \multicolumn{1}{c|}{87.4} & 85.8 & \multicolumn{1}{c|}{85.6} & 87.3 & \multicolumn{1}{c|}{87.3} & 80.7 & \multicolumn{1}{c|}{80.6} & 85.8 & \multicolumn{1}{c|}{85.6} & 81.7 & \multicolumn{1}{c|}{81.7} & 90.9 & \multicolumn{1}{c|}{90.3} & 85.7 & 85.5 \\ \bottomrule  
\end{tabular} 
} \vspace{-0.25em}
\caption{\textbf{Ablation study on DomainNet} between co-training and one-direction teaching under 3-shot settings. } \vspace{-0.85em}\label{ablation_study_cotraining}
\end{table*}

\subsection{Comparison Results} 
We evaluate our ECB method and the previous SOTA methods on \textit{Office-Home} and \textit{DomainNet} under UDA and SSDA settings, respectively. For a fair comparison with other DA methods, we rely on classification outcomes from the CNN branch, employing ResNet-50 as the backbone for UDA and ResNet-34 for SSDA. As a result, our method does not add any testing complexity compared to previous DA approaches. \textit{Besides, we have included further details about the experimental results in Supplementary Materials}.

\noindent\textbf{Results on \textit{Office-Home} under UDA setting.} The implementation of our ECB method has significantly boosted classification efficiency in all domain shift tasks, consistently outperforming the comparison methods, as detailed in \cref{result_officehome_uda}. Remarkably, our approach has recorded accuracy enhancements of +7.7\%, +8.1\%, and +7.2\% for the $C{\rightarrow}A$, $C{\rightarrow}R$, and $P{\rightarrow}A$ tasks, respectively, surpassing the results of the second-best. In addition, our method has achieved an impressive average classification accuracy of 81.2\%, showing a remarkable margin of +5.4\% over the nearest-competitor EIDCo \cite{EIDCo}.

\noindent\textbf{Results on \textit{DomainNet} under SSDA setting.} The results on the \textit{DomainNet} dataset are presented for both 1-shot and 3-shot settings in \cref{result_domainnet}, where the CNN branch of our ECB method outperforms all prior methods. In comparison to the nearest-competitor method, G-ABC \cite{G-ABC}, the ECB (CNN) achieves an impressive maximum performance increase of +9.3\% in the $skt{\rightarrow}pnt$ task for 3-shot learning. Even in the more restrictive 1-shot learning, the ECB method demonstrates robust performance, showing a minimum increase of +3.1\% in the $rel{\rightarrow}clp$ task. On average, the ECB method validates a performance improvement of +6.6\% in the 1-shot setting and +7.1\% in the 3-shot setting. 
  

\begin{figure}[!t]
\centering
\begin{subfigure}[t]{0.49\linewidth}
\includegraphics[width=115pt]{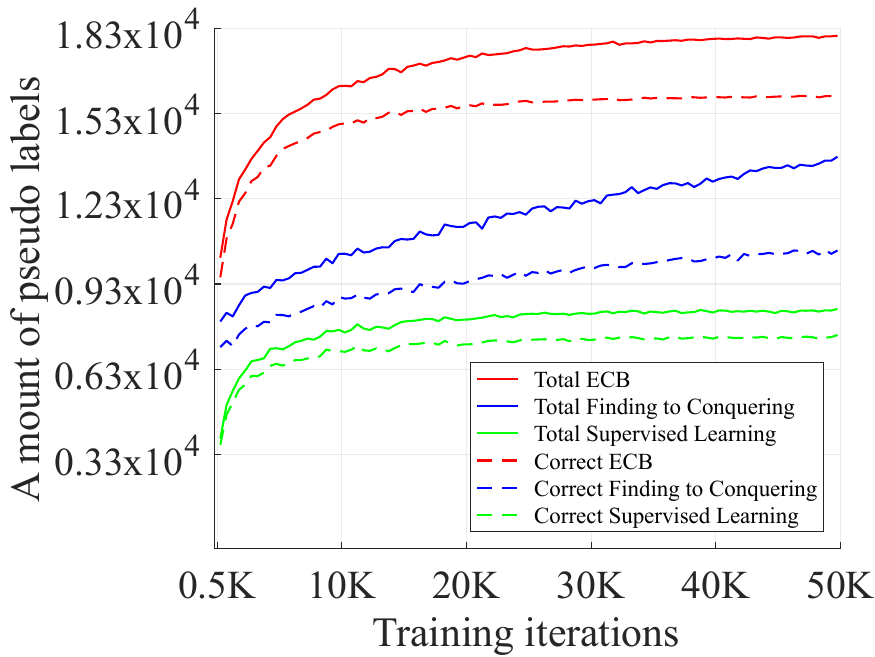}
\caption{}\label{sub:4a}
\end{subfigure}
\begin{subfigure}[t]{0.49\linewidth}
\includegraphics[width=115pt]{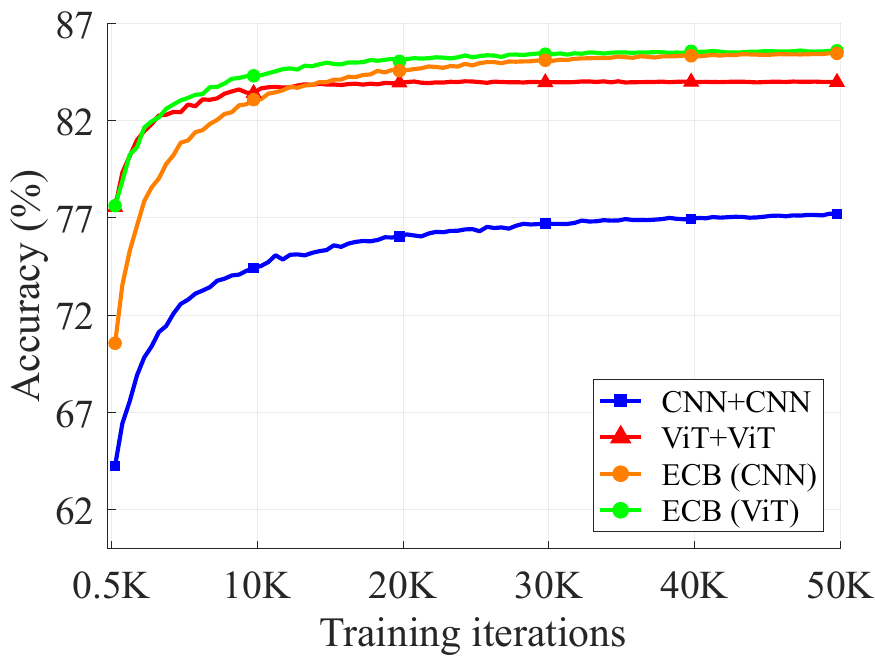}
\caption{}\label{sub:4b}
\end{subfigure} \vspace{-0.5em}
\caption{(a) The quality and quantity of the pseudo labels are generated by the CNN branch on \textit{DomainNet} under the 3-shot setting of the $rel{\rightarrow}clp$ task using ResNet-34. (b) Comparison between backbone settings on \textit{DomainNet} under the 3-shot setting. Displayed is the mean accuracy across all domain shift tasks.} \vspace{-1.6em} \label{fig_4} 
\end{figure}

\section{Analyses} \label{sec:Analyses} 
\textbf{Ablation studies.} A single classifier is concurrently trained with both labeled source and target data. It is thus easily dominated by the labeled source data. This imbalance prevents the classifier from using the extra-labeled target data effectively, resulting in a misalignment between the learned and true class-specific boundaries. This misalignment is called data bias. Therefore, we perform an ablation study to evaluate the effectiveness of each stage in the proposed method. The emphasis is on addressing the data bias issue within the ViT branch, leveraging the CNN branch's capability to generate high-quality pseudo labels during the $rel{\rightarrow}clp$ task on \textit{DomainNet} under, as illustrated in \cref{sub:4a}. Initially, during the Supervised Training phase, we are only able to generate around 8,000 total pseudo labels. Following the integration of the FTC strategy, the total pseudo labels smoothly increase up to 14,000 with steady maintenance of 10,000 correct pseudo labels. This plays a crucial role in substantially mitigating data bias by generating numerous pseudo-labels, which improves the diversity and representativeness of the unlabeled target dataset. The final experiment is implemented to highlight that co-training is necessary for reducing the remaining gap following the FTC stage, which yields a total of 18,000 pseudo labels and 16,000 correct pseudo labels. 

\begin{table}[!t]
\resizebox{\columnwidth}{!}{%
\centering
    \begin{tabular}{c|cc|cc} \toprule
     \textbf{Cannon} & \multicolumn{1}{c|}{CNN (Before)} & ViT (Before) & \multicolumn{1}{c|}{CNN (After)} &  ViT (After) \\ \midrule
     \multicolumn{1}{c|}{\includegraphics[width=0.3\linewidth]{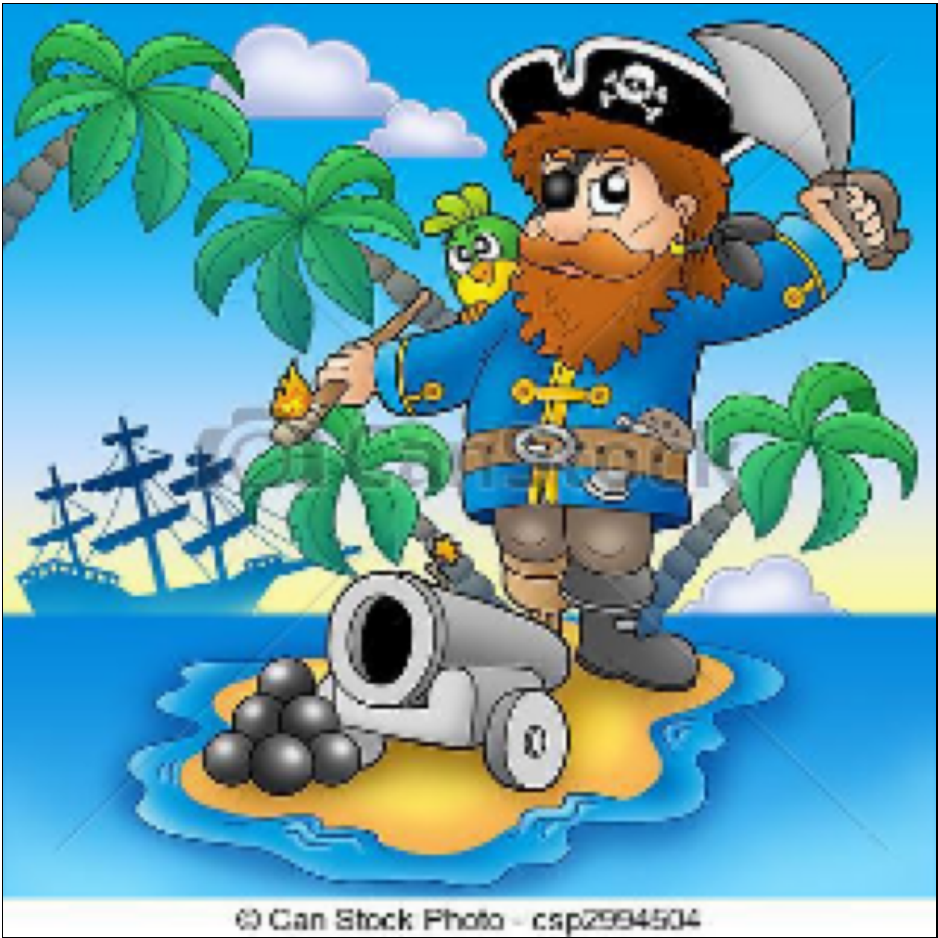}} & \multicolumn{1}{c|}{\includegraphics[width=0.3\linewidth]{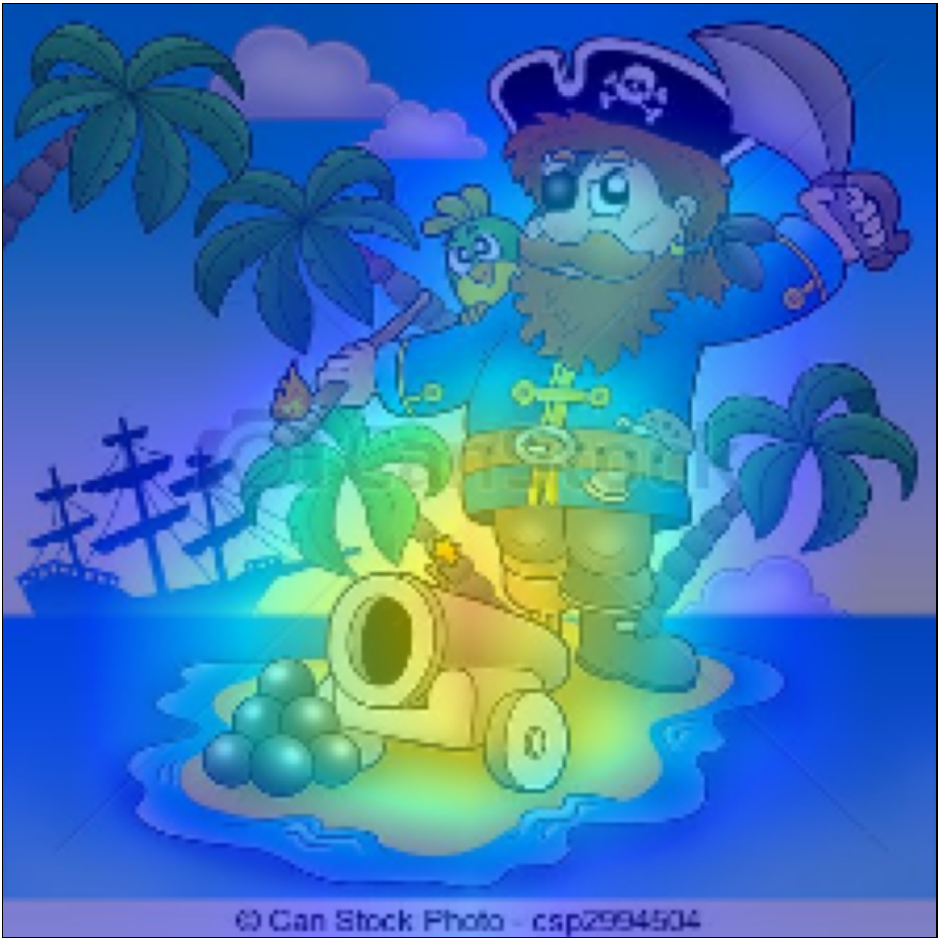}} & \multicolumn{1}{c|}{\includegraphics[width=0.3\linewidth]{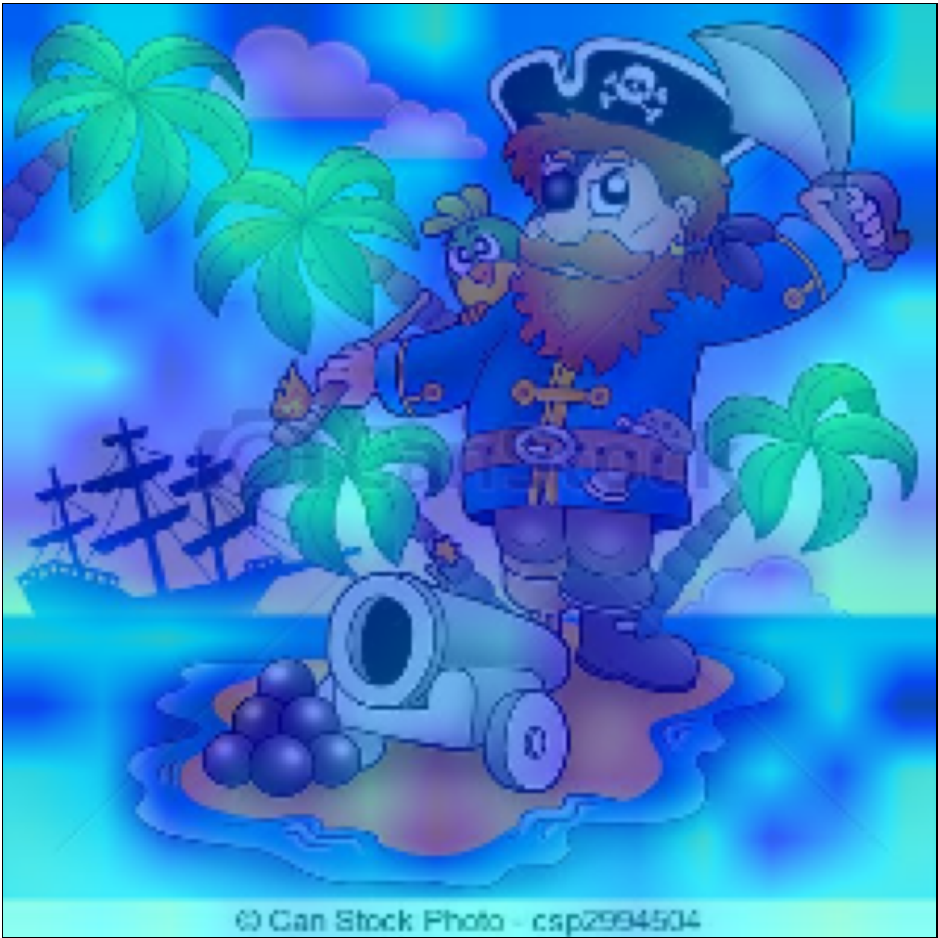}} &  \multicolumn{1}{c|}{\includegraphics[width=0.3\linewidth]{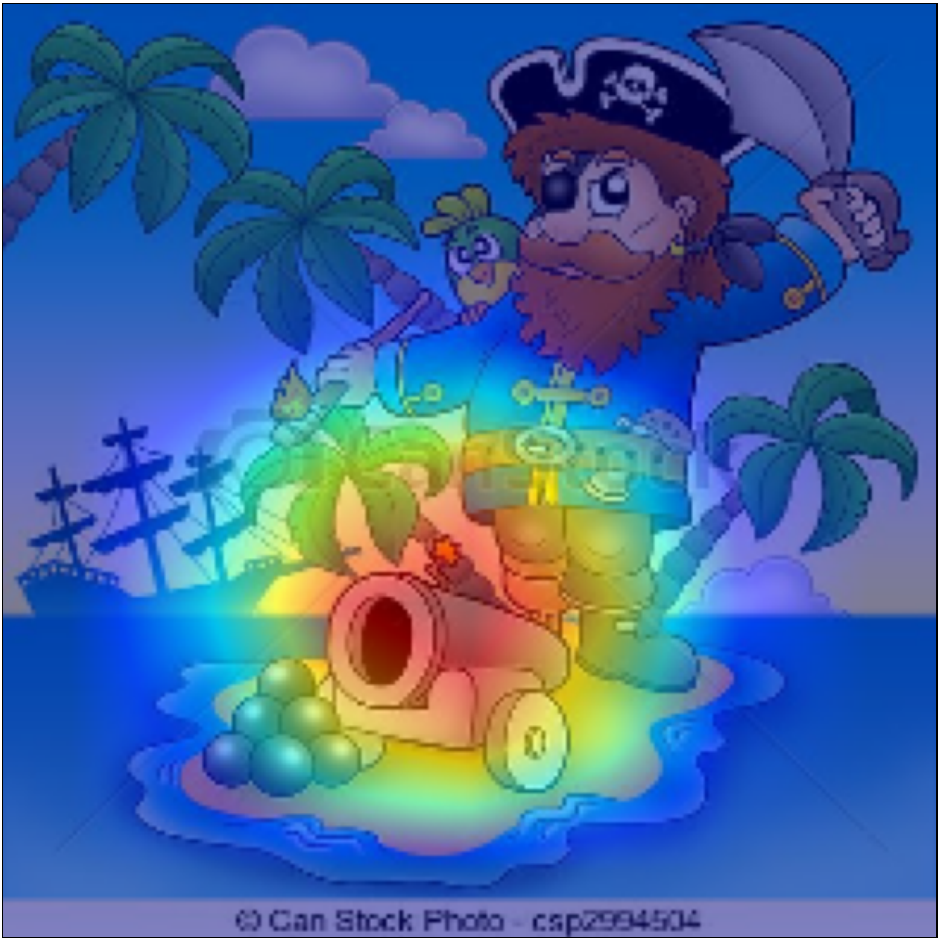}} & \multicolumn{1}{c}{\includegraphics[width=0.3\linewidth]{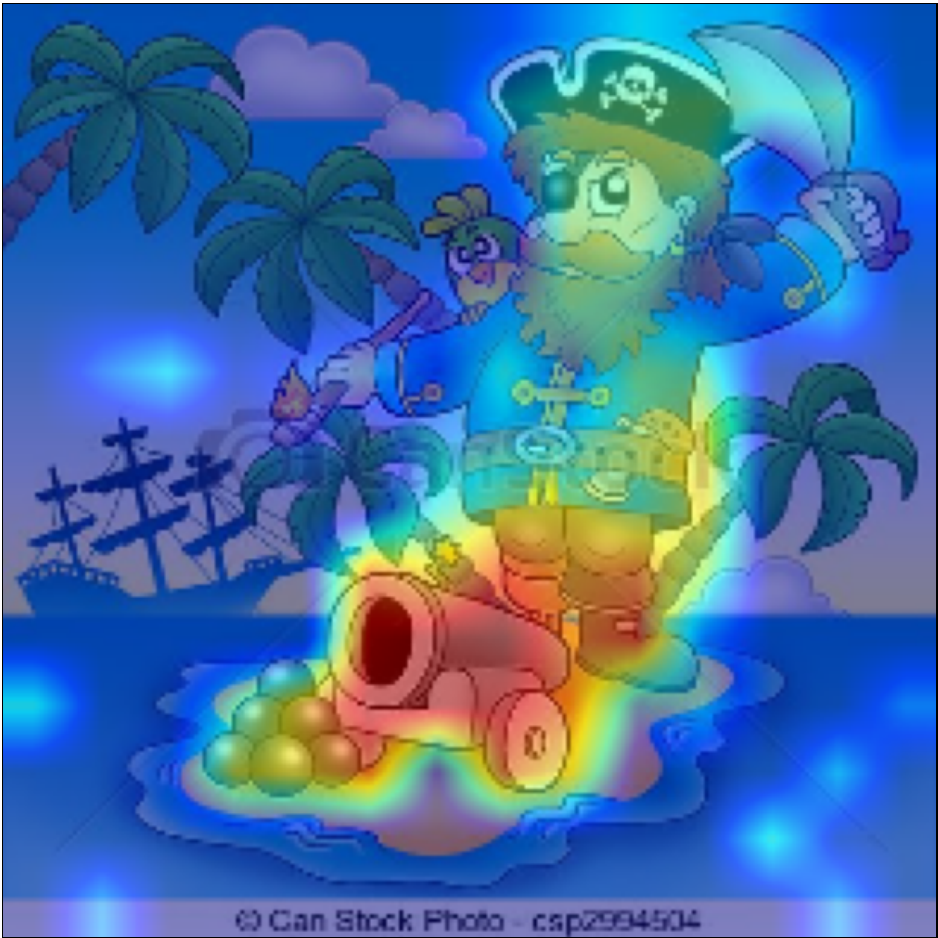}} \\ \midrule
     \textbf{Bird} & \multicolumn{1}{c|}{CNN (Before)} & ViT (Before) & \multicolumn{1}{c|}{CNN (After)} &  ViT (After) \\ \midrule
     \multicolumn{1}{c|}{\includegraphics[width=0.3\linewidth]{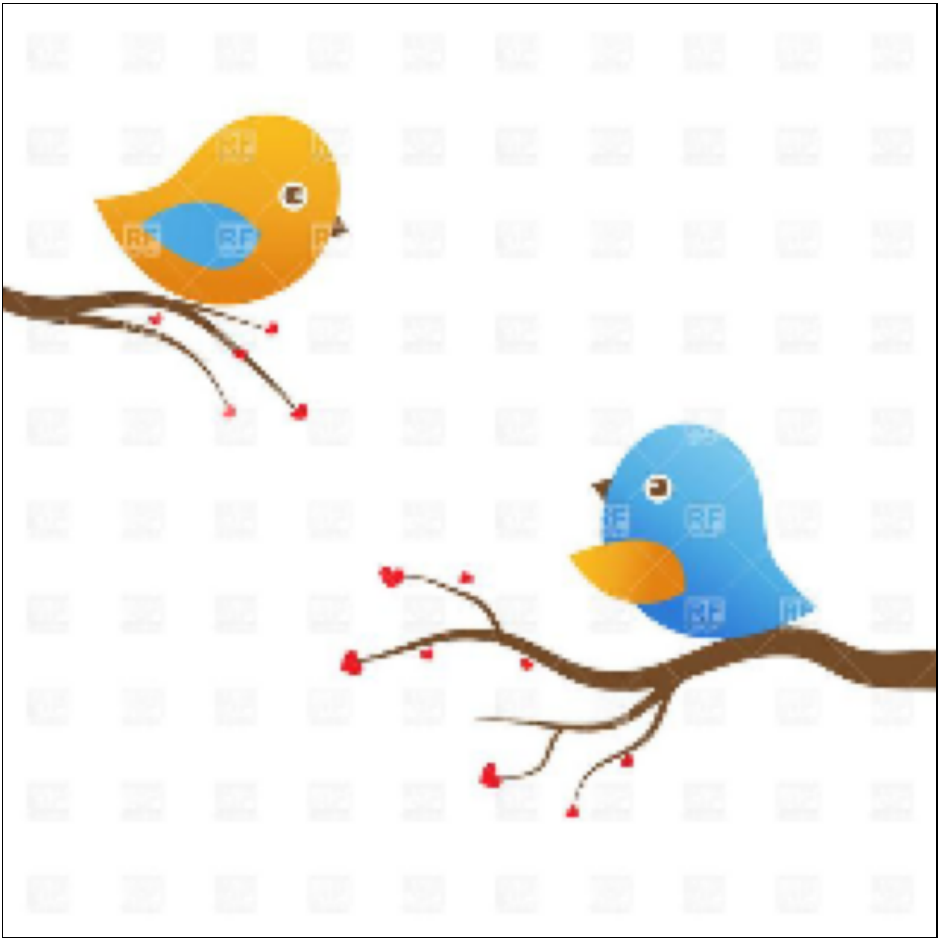}} & \multicolumn{1}{c|}{\includegraphics[width=0.3\linewidth]{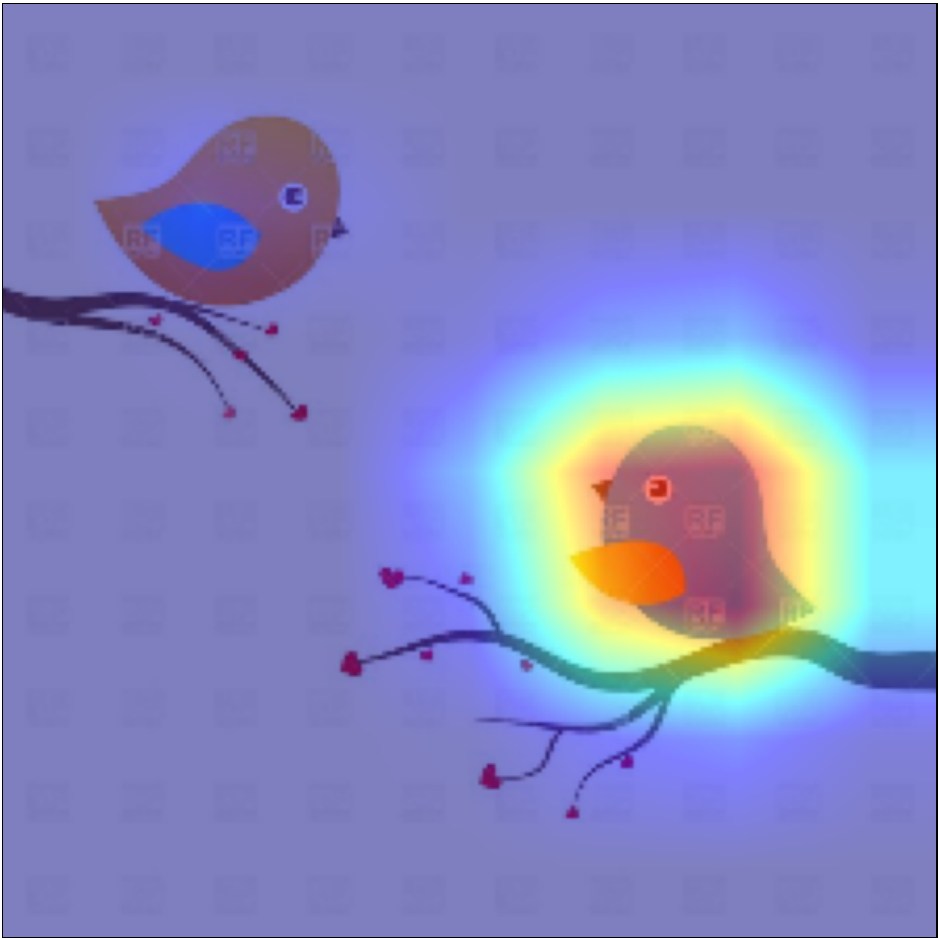}} & \multicolumn{1}{c|}{\includegraphics[width=0.3\linewidth]{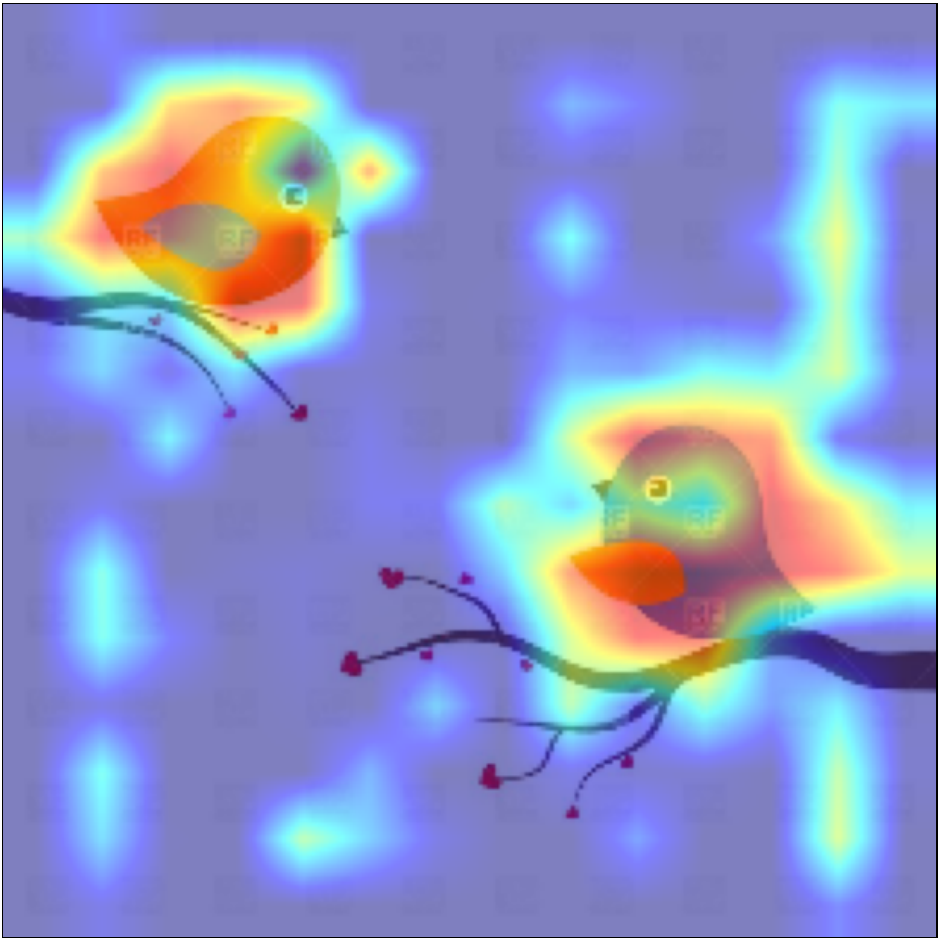}} & \multicolumn{1}{c|}{\includegraphics[width=0.3\linewidth]{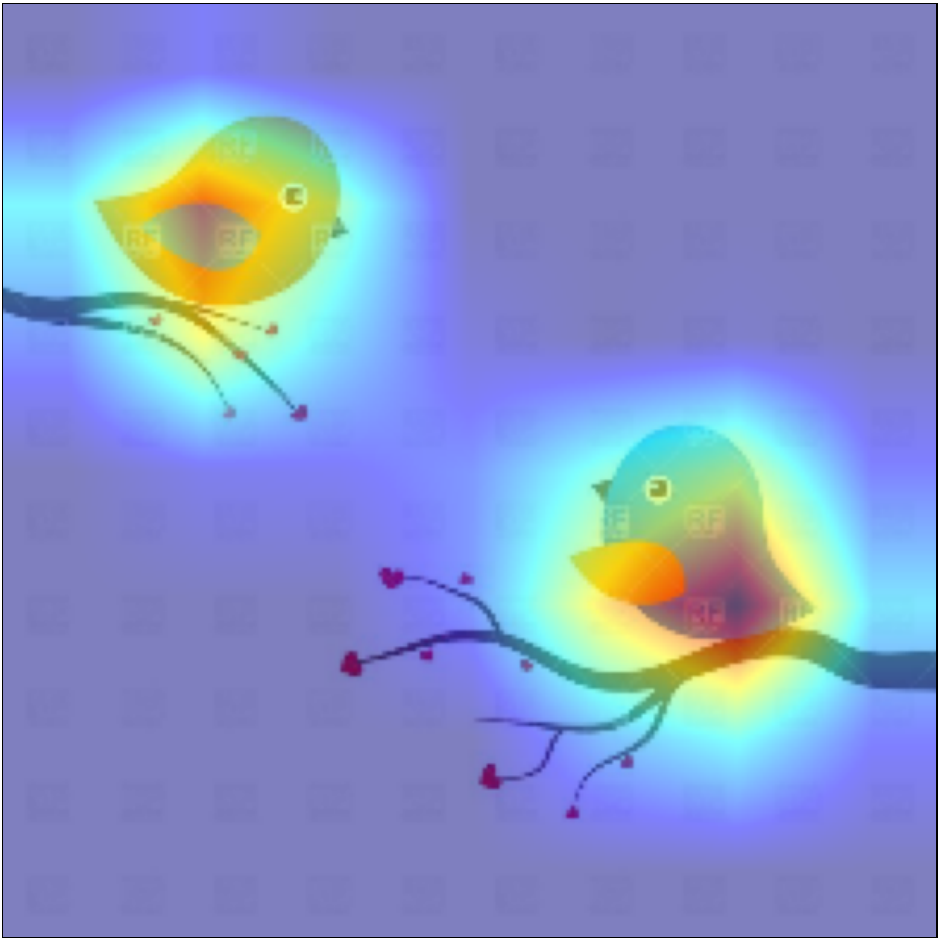}}  & \multicolumn{1}{c}{\includegraphics[width=0.3\linewidth]{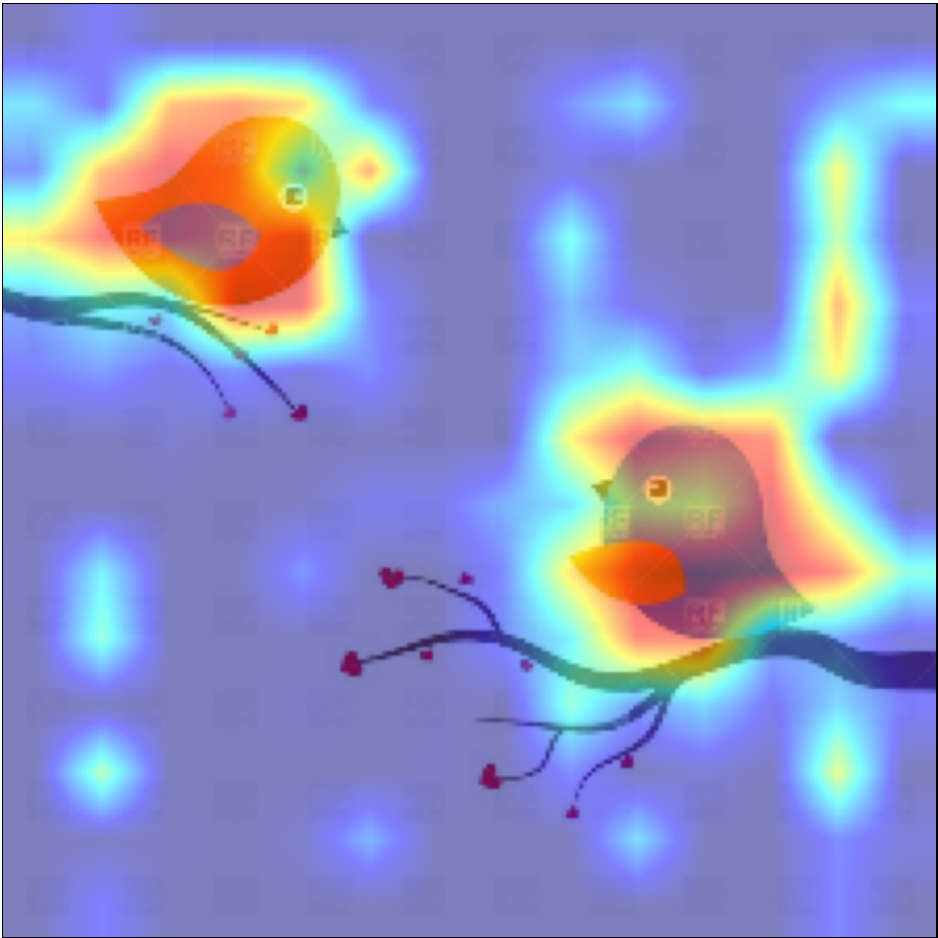}} \\
    \bottomrule
    \end{tabular} 
} \vspace{-0.2em}
\caption{\textbf{Visualize the feature maps} for the `\textit{Cannon}' and `\textit{Bird}' examples to investigate the learning behaviors of CNN and ViT with and without using the proposed method ECB.} \label{grad_cam} \vspace{-1.31em}
\end{table}




\begin{figure*}[!t]
    \centering
    \begin{subfigure}[t]{0.24\linewidth}
        \includegraphics[width=110pt]{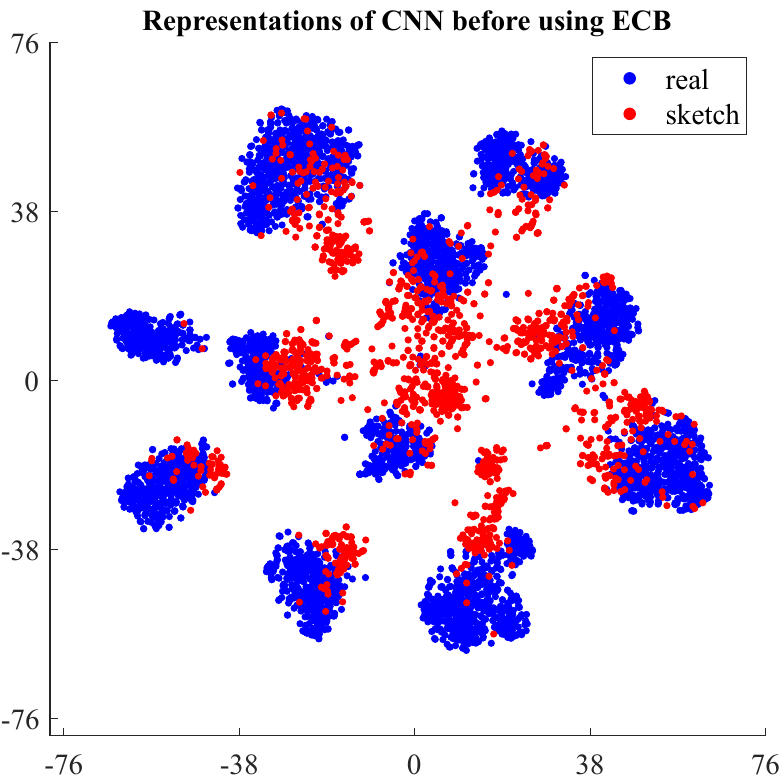}
        \caption{}
        \label{visualize_tsne_a}
    \end{subfigure}
    \begin{subfigure}[t]{0.24\textwidth}
        \includegraphics[width=110pt]{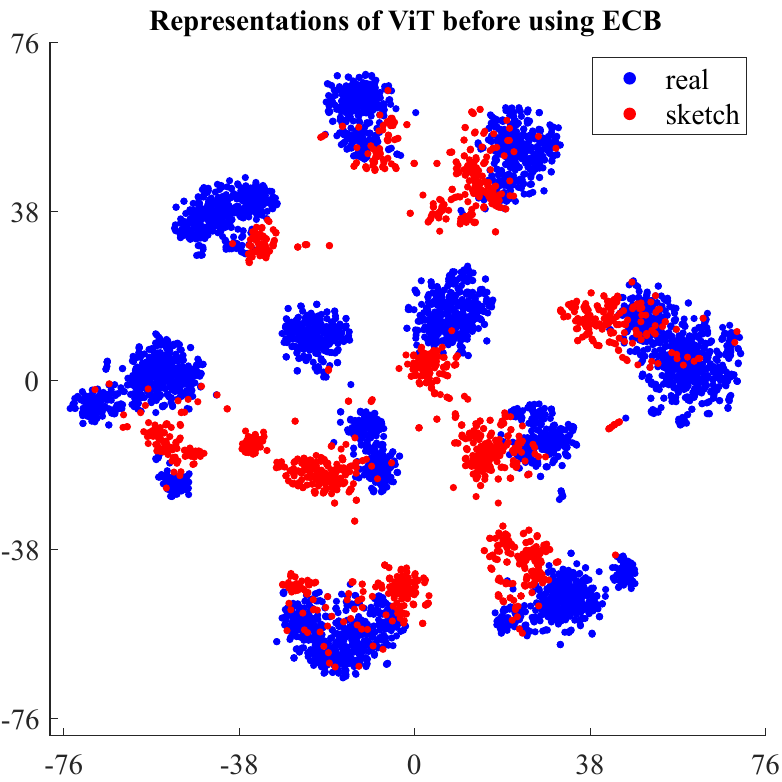}
        \caption{}
        \label{visualize_tsne_b}
    \end{subfigure}
    \begin{subfigure}[t]{0.24\textwidth}
        \includegraphics[width=110pt]{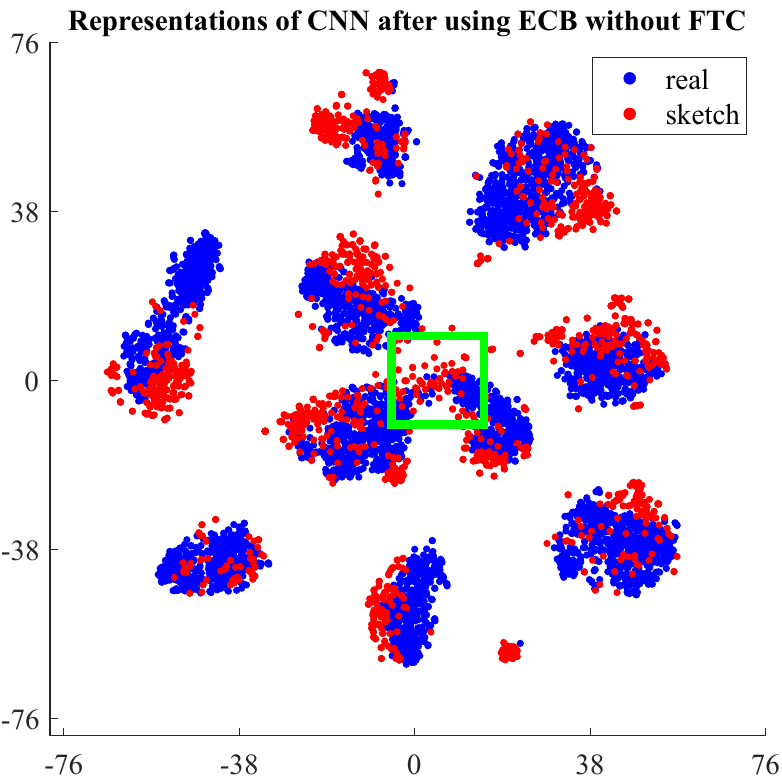}
        \caption{}
        \label{visualize_tsne_c}
    \end{subfigure}
    \begin{subfigure}[t]{0.24\textwidth}
        \includegraphics[width=110pt]{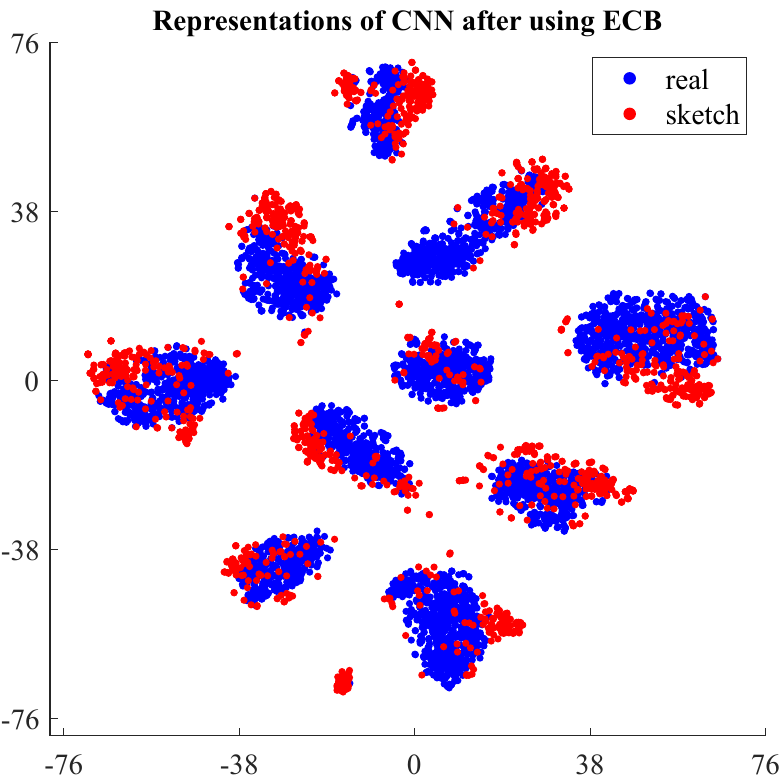}
        \caption{}
        \label{visualize_tsne_d}
    \end{subfigure} \vspace{-0.75em}
    \caption{We visualize feature spaces for the $rel{\rightarrow}skt$ task on \textit{DomainNet} in the 3-shot scenario using t-SNE \cite{tsne}. Figures (a) and (b) illustrate the features obtained by CNN and ViT branches before adaptation, respectively. Figures (c) and (d) showcase the distribution changes of the CNN branch depending on the presence of the FTC strategy when implementing our ECB method.} \vspace{-1.0em}
    \label{visualize_tsne} 
\end{figure*}

\noindent\textbf{Effectiveness of co-training.} We further investigate a variant of ECB termed one-direction teaching on the \textit{DomainNet} under the 3-shot setting. In this approach, we employ either $\mathcal{L}^{unl}_{{vit}\rightarrow{cnn}}$ or $\mathcal{L}^{unl}_{{cnn}\rightarrow{vit}}$ to generate pseudo labels for the remaining branch, while maintaining the standard configurations through the use of Supervised Training followed by FTC strategy. As depicted in \cref{ablation_study_cotraining}, there is a performance drop when using one-direction teaching. Specifically, in the $vit{\rightarrow}cnn$ scenario, the ViT branch generates pseudo labels for the unlabeled target data to teach the CNN branch through minimizing $\mathcal{L}^{unl}_{{vit}\rightarrow{cnn}}$. It is crucial to highlight that the ViT branch does not receive pseudo labels from the CNN branch, resulting in the CNN branch outperforming the ViT branch by +3.3\%. In contrast, in the $cnn{\rightarrow}vit$ scenario, where $\mathcal{L}^{unl}_{{cnn}\rightarrow{vit}}$ is minimized, the average performance on the ViT branch is +10.9\% more outstanding than the CNN branch. However, it is worth mentioning that the performance on the ViT branch suffers from -1.1\% reduction over the $vit{\rightarrow}cnn$ scenario. This performance drop is attributed to the introduction of noise by the CNN branch during the learning process of the ViT branch. On the other hand, when employing $\mathcal{L}^{unl}_{{vit}\rightarrow{cnn}}$, the CNN branch provides a significant boost in the average performance by +15.3\%, highlighting the superior generalization capabilities of the ViT branch compared to the CNN branch. These scenarios try to leverage the unique strengths of CNN and ViT through the co-training strategy, showcasing the mutual exchange of knowledge between two branches and its potential for generalizing to unlabeled target data. As a result of co-training, the average performance on the target domain of the ViT and CNN branches reaches the optimal results of 85.7\% and 85.5\%, respectively. 

\noindent\textbf{Architecture analysis.} We examine the effectiveness of the unique strengths of ViT and CNN over variations of architectures such as \say{CNN + CNN} and \say{ViT + ViT} on \textit{DomainNet} under 3-shot settings using ResNet-34 for CNN and ViT/B-16 for ViT as shown in \cref{sub:4b}. On average, the results show that \say{CNN + CNN} only achieves around 77.0\% due to the lack of global information when just using CNN. In addition, the introduction of ViT with the primary goal of improving performance drives the \say{ViT + ViT} architecture to achieve a significant increase of +7.0\% compared to the \say{CNN + CNN} architecture. This architecture makes it possible to establish more general class-specific boundaries, yet is unfair when compared with the previous SSDA methods \cite{MME, MCD, ATDOC, DECOTA, SLA} using ResNet-34 as the backbone, and the lack of local representation is extremely important. As a result, the evidence shows that combining \say{CNN + ViT} with the bridge of co-training achieves the highest accuracy of about 85.5\% for both branches.

\noindent\textbf{Attention map visualization analysis.}
To demonstrate the effectiveness of our designing framework methodology, we provide the attention map visualization results by using Grad-CAM \cite{grabcam}, as visualized in \cref{grad_cam}.  \textit{CNN supports ViT}: Before applying ECB, ViT was background-sensitive in the ``\textit{Cannon}" class, whereas CNN still captures the correct object. However, after applying ECB, ViT can explicitly recognize the target object and back to enhance the robustness of CNN. \textit{ViT supports CNN}: In the ``\textit{Bird}" class before applying ECB, CNN only obtains the part of the target, while ViT can cover the whole input object. Then, CNN is complemented to focus on the object accurately by ViT using ECB. Consequently, these findings validate that both branches offer distinct expertise and enhance each other rather than one overshadowing the other.

\noindent\textbf{Feature Visualization.} We present a detailed visualization of the feature space for the \textit{DomainNet} dataset within the $rel{\rightarrow}skt$ task under the 3-shot setting. This visualization distinctly highlights the source domain (blue-colored) and the target domain (red-colored). \cref{visualize_tsne_a,visualize_tsne_b} show the domain alignment of both the CNN and ViT branches before adaptation. In particular, \cref{visualize_tsne_a} visualizes the feature space extracted by the CNN branch before applying ECB, reflecting a scatter that indicates weak classifier performance. In contrast, \cref{visualize_tsne_b} displays the well-defined clusters in the ViT branch for the same unlabeled target data, which emphasizes the robustness of ViT in identifying more general class-specific boundaries compared to the CNN branch. \cref{visualize_tsne_c,visualize_tsne_d} show the distribution changes of the CNN branch depending on the presence of the FTC strategy when implementing ECB. Initially, \cref{visualize_tsne_c} indicates that the target representations overlap (highlighted with green box) when implementing ECB without the FTC strategy. On the other hand, \cref{visualize_tsne_d} shows the well-aligned source and target domain representations with clusters to distinct separation when applying the FTC strategy, which demonstrates the effectiveness of our proposed method. 
\section{Conclusion} \label{sec:conclusion}
In this work, we have developed a novel method for learning CNN on ViT with ECB strategy, taking advantage of ViT and CNN. This innovative approach focuses on reducing data bias and significantly improves the precision of pseudo labels generated, which aligns between source and target domains. Our method is also fair when evaluated on the CNN branch and outperforms the previous SOTA methods on various DA benchmark datasets.

\noindent\textbf{Discussions.}\quad   
We found that identifying an optimal threshold pair \{$\tau_{vit}$ and $\tau_{cnn}$\} was quite time-consuming. Therefore, we leave an open task for future research that uses a dynamic threshold algorithm for domain adaptation instead of the fixed threshold.

\noindent\textbf{Acknowledgement.} This work was supported by the National Research Foundation of Korea (NRF) grant funded by the Korea Government (MSIT) (No. RS-2023-00214326 and No. RS-2023-00242528).

{
    \small
    \bibliographystyle{ieeenat_fullname}
    \bibliography{main}
}
\end{document}


\maketitle
In this Supplementary Materials, we provide expanded details on our analyses and additional experimental results. This section encompasses four key appendices: \textcolor{red}{Appendix 1} delves into the notations used throughout our study. \textcolor{red}{Appendix 2} describes the algorithm we adopted. \textcolor{red}{Appendix 3} shows the sensitivity of two fixed thresholds, specifically $\tau_{vit}$ and $\tau_{cnn}$, providing an in-depth analysis of their impact. \textcolor{red}{Appendix 4} presents additional experiment results.

\section{Notations}
We present the notations commonly employed in our method, as outlined in {\cref{notation}}.

\begin{table}[t]
\begin{adjustbox}{width=\columnwidth,center}
\begin{tabular}{lp{8.2cm}}
\toprule
        \textbf{Notations}                & \textbf{Descriptions} \\ \midrule 
        $\mathcal{D_S}$                   & The set of source samples. \\ \midrule
        $x^{\mathcal{S}}_i$                           & The {\textit{i}-th} sample in the source domain. \\ \midrule
        $y^{\mathcal{S}}_i$                           & The label of the \textit{i}-th sample in the source domain. \\ \midrule
        $\mathcal{N_S}$                   & The number of source samples. \\ \midrule
        $\mathcal{D}_{\mathcal{T}_l}$     & The set of labeled target samples. \\ \midrule
        $x^{\mathcal{T}_l}_i$             & The {\textit{i}}-th sample in the labeled target domain. \\ \midrule
        $y^{\mathcal{T}_l}_i$             & The label of the {\textit{i}}-th sample in the labeled target domain. \\ \midrule
        $\mathcal{N}_{\mathcal{T}_l}$     & The number of labeled target samples. \\ \midrule
        $\mathcal{D}_{\mathcal{T}_u}$     & The set of unlabeled target samples. \\ \midrule
        $x^{\mathcal{T}_u}_i$             & The {\textit{i}}-th unlabeled target sample in the target domain. \\ \midrule
        $y^{\mathcal{T}_u}_i$             & The label of the {\textit{i}}-th sample in the unlabeled target domain. \\ \midrule
        $\mathcal{N}_{\mathcal{T}_u}$     & The number of unlabeled target samples. \\ \midrule
        $\mathcal{D}_l$                 & The set of labeled samples. \\ \midrule
        $x^l_i$             & The {\textit{i}}-th labeled sample. \\ \midrule
        $y^l_i$             & The label of the {\textit{i}}-th labeled sample. \\ \midrule
        $\mathcal{N}_l$     & The number of labeled samples. \\ \midrule
        $Aug_{w}(\cdot)$                  & The weak augmentation.   \\ \midrule
        $Aug_{str}(\cdot)$                & The strong augmentation.  \\ \midrule
        $x_{i,w}^{\mathcal{T}_u}$         & The weakly augmented $i$-th unlabeled target sample. \\ \midrule
        $x_{i,str}^{\mathcal{T}_u}$       & The strongly augmented $i$-th unlabeled target sample. \\ \midrule
        ${E_1}( \cdot ;{\boldsymbol{\theta} _{{E_1}}})$    & The ViT encoder. \\ \midrule
        ${E_2}( \cdot ;{\boldsymbol{\theta} _{{E_2}}})$    & The CNN encoder. \\ \midrule
        ${F_1}( \cdot ;{\boldsymbol{\theta} _{F_1}})$      & The classifier of ViT branch. \\ \midrule
        ${F_2}( \cdot ;{\boldsymbol{\theta} _{F_2}})$      & The classifier of CNN branch. \\ \midrule
        $p^{l}_{1}(x^{l}_{i})$                       & The ViT branch's probability on the labeled sample $x^{l}_{i}$. \\ \midrule
        $p^{l}_{2}(x^{l}_{i})$                       & The CNN branch's probability on the labeled sample $x^{l}_{i}$. \\ \midrule
        $p^{find}_{1}(x^{\mathcal{T}_u}_{i})$                       & The probability output of $F_1$ with ViT encoder $E_1$ on the unlabeled target sample, $x^{\mathcal{T}_u}_{i}$. \\ \midrule
        $p^{find}_{2}(x^{\mathcal{T}_u}_{i})$                       & The probability output of $F_2$ with ViT encoder $E_1$ on the unlabeled target sample, $x^{\mathcal{T}_u}_{i}$. \\ \midrule
        $p^{conq}_{1}(x^{\mathcal{T}_u}_{i})$                       & The probability output of $F_1$ with CNN encoder $E_2$ on the unlabeled target sample, $x^{\mathcal{T}_u}_{i}$. \\ \midrule
        $p^{conq}_{2}(x^{\mathcal{T}_u}_{i})$                       & The probability output of $F_2$ with CNN encoder $E_2$ on the unlabeled target sample, $x^{\mathcal{T}_u}_{i}$. \\ \midrule
        $\tau_{vit}$           & The fixed threshold of $vit{\rightarrow}cnn$. \\ \midrule
        $\tau_{cnn}$           & The fixed threshold of $cnn{\rightarrow}vit$. \\ \midrule
        $\hat{q}^{v}_{i}$                 & The pseudo label generated by the ViT branch on the weakly unlabeled target sample $x^{\mathcal{T}_u}_{i,w}$. \\ \midrule
        $p^{c}(x^{\mathcal{T}_u}_{i,str})$ & The CNN branch's probability on the strongly unlabeled target sample  $x^{\mathcal{T}_u}_{i,str}$. \\ \midrule
        $\hat{q}^{c}_{i}$                 & The pseudo label generated by the CNN branch on the weakly unlabeled target sample $x^{\mathcal{T}_u}_{i,str}$. \\ \midrule
        $p^{v}(x^{\mathcal{T}_u}_{i,str})$& The ViT branch's probability on the strongly unlabeled target sample  $x^{\mathcal{T}_u}_{i,str}$. \\
        \bottomrule
\end{tabular} \vspace{-0.25em}
\end{adjustbox} 
\caption{The notations commonly employed in our method.} \vspace{-0.1em} \label{notation}
\end{table}
\section{Algorithm}
In summary, the whole algorithm to train the proposed ECB is shown in \cref{algorithm_1}.

\begin{algorithm}
    \caption{The ECB algorithm} \label{alg_1}
    \begin{algorithmic}[1] 
    
    \STATE \textbf{Data setting:} 

    \begin{itemize}
        \item The labeled source data  $\mathcal{D_S} = \{(x^{\mathcal{S}}_i, y^{\mathcal{S}}_i)\}^{\mathcal{N_S}}_{i=1}$.
        \item The labeled target data $\mathcal{D}_{\mathcal{T}_l} = \{{(x^{\mathcal{T}_l}_i, y^{\mathcal{T}_l}_i)\}}^{\mathcal{N}_{\mathcal{T}_l}}_{i=1}$. \textit{Notably, $\mathcal{D}_{\mathcal{T}_l}$ is empty in UDA scenario}.
        \item The unlabeled target data $\mathcal{D}_{\mathcal{T}_u} = \{{(x^{\mathcal{T}_u}_i, y^{\mathcal{T}_u}_i)\}}^{\mathcal{N}_{\mathcal{T}_u}}_{i=1}$. 
    \end{itemize}
    
    \noindent \textcolor{red}{Note}: The labeled data $\mathcal{D}_l=\mathcal{D_S} \cup \mathcal{D}_{\mathcal{T}_l}$. \vspace{0.5em}
    
    \STATE \textbf{Architectures:}
    
    \noindent The \textbf{ViT branch}: a ViT encoder ${E_1}( \cdot ;{\boldsymbol{\theta}_{{E_1}}})$ and a classifier ${F_1}( \cdot ;{\boldsymbol{\theta}_{{F_1}}})$.
    
    \noindent The \textbf{CNN branch}: a CNN encoder ${E_2}( \cdot ;{\boldsymbol{\theta}_{{E_2}}})$ and a classifier ${F_2}( \cdot ;{\boldsymbol{\theta}_{{F_2}}})$. \vspace{0.5em}

    \STATE \textbf{Hyperparameters:} Fixed thresholds $\tau_{vit}$ and $\tau_{cnn}$, the number of training interations $T$, learning rates for ViT and CNN, $\eta_{vit}$ and $\eta_{cnn}$. \vspace{0.5em}

    \STATE \textbf{Traning strategy:}
    
    \FOR {$t {\leftarrow} 1$ to $T$}
        \STATE {\textcolor{orange}{\# Supervised Training}} 

         $\boldsymbol{\theta}_{{E_1}}$, $\boldsymbol{\theta}_{{F_1}}$ $\leftarrow$ $\eta_{vit}$$\nabla \mathcal{L}^{sup}_{vit}$; \vspace{0.25em}

         $\boldsymbol{\theta}_{{E_2}}$, $\boldsymbol{\theta}_{{F_2}}$ $\leftarrow$ $\eta_{cnn}$$\nabla \mathcal{L}^{sup}_{cnn}$;
        
        \STATE {\textcolor{orange}{\# Finding to Conquering}} 
        \STATE {\textcolor{blue}{{$\rhd$} \textit{Finding Stage}}}

        $\boldsymbol{\theta}_{{F_1}}$, $\boldsymbol{\theta}_{{F_2}}$ $\leftarrow$ $\eta_{vit}$$\nabla \mathcal{L}_{find}$;
            
        \STATE {\textcolor{blue}{{$\rhd$} \textit{Conquering Stage}}}
         
        $\boldsymbol{\theta}_{{E_2}}$ $\leftarrow$ $\eta_{cnn}$$\nabla \mathcal{L}_{conq}$;

        \STATE {\textcolor{orange}{\# Co-training}}
        
        \STATE  {\textcolor{blue}{{$\rhd$} \textit{The ViT branch teaches the CNN branch}}}

        $\boldsymbol{\theta}_{{E_2}}$, $\boldsymbol{\theta}_{{F_2}}$ $\leftarrow$ $\eta_{cnn}$ $\nabla \mathcal{L}^{unl}_{{vit}\rightarrow{cnn}}$;
        
        \STATE  {\textcolor{blue}{{$\rhd$} \textit{The CNN branch teaches the ViT branch}}}

        $\boldsymbol{\theta}_{{E_1}}$, $\boldsymbol{\theta}_{{F_1}}$ $\leftarrow$ $\eta_{vit}$ $\nabla \mathcal{L}^{unl}_{{cnn}\rightarrow{vit}}$;    
    \ENDFOR \vspace{0.5em}

    \STATE \textbf{Inference:} $\hat{y}^{\mathcal{T}_u}_{i} = \argmax \big(F_{2}(E_{2}(x^{\mathcal{T}_u}_{i}))\big)$. 
    \end{algorithmic} \label{algorithm_1} 
\end{algorithm}

\section{Sensitivity of threshold $\tau_{vit}$ and $\tau_{cnn}$}
\begin{figure}[!t]
\centering
\includegraphics[width=225pt]{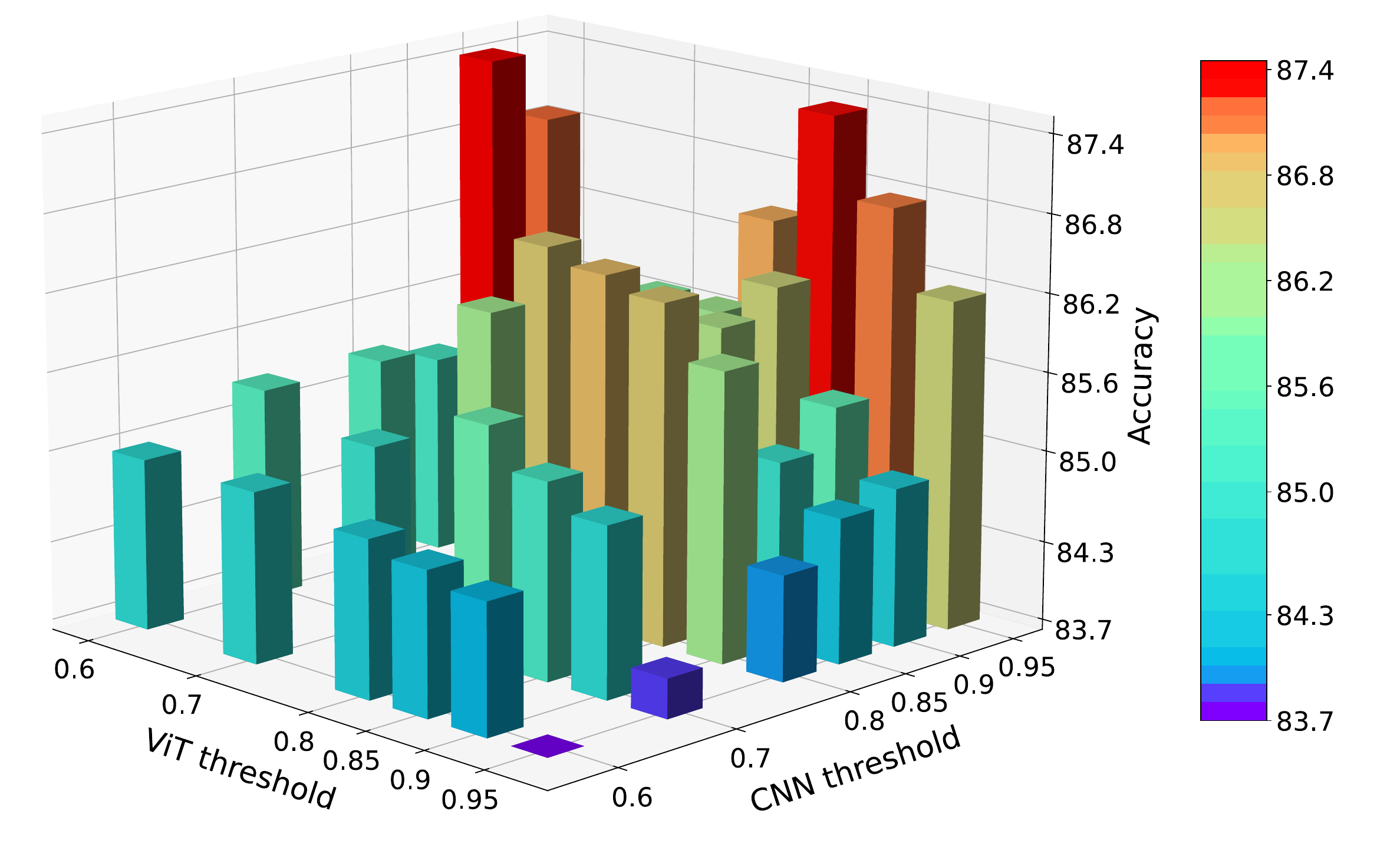}
\caption{We evaluated our method's performance on the CNN branch by adjusting $\tau_{vit}$ and $\tau_{cnn}$ to the values \{$0.6$, $0.7$, $0.8$, $0.85$, $0.9$ and $0.95$\}. All experiments were conducted in the SSDA scenario under 3-shot setting of $rel{\rightarrow}clp$ task.}  \label{histogram_thresh} 
\end{figure}

In the semi-supervised domain adaptation (SSDA) scenario, we assess the performance sensitivity of our method on the CNN branch by varying the threshold values $\tau_{vit}$ and $\tau_{cnn}$ in the $rel{\rightarrow}clp$ scenario under 3-shot setting on \textit{DomainNet} \cite{DomainNet}. Specifically, we select threshold values of \{0.6, 0.7, 0.8, 0.85, 0.9, and 0.95\} for our analysis as shown in \cref{histogram_thresh}. In total, $36$ experiments are conducted to gauge the impact of these thresholds on our approach. Notably, the optimal performance of 87.4\% is attained when $\tau_{vit} = 0.6$ and $\tau_{cnn} = 0.9$. This suggests that ViT's predictions tend to generate pseudo labels more confidently than those of CNN during the training phase. Thus, a lower ViT threshold ($\tau_{vit} = 0.6$) not only boosts the number of pseudo labels but also produces reliable ones to improve the guidance of the CNN branch more effectively. Conversely, the CNN branch shows less certainty with unlabeled target samples, requiring a higher threshold ($\tau_{cnn} = 0.9$). Furthermore, in the case $\tau_{vit}=0.85$ and $\tau_{cnn}=0.95$, we observe considerable performance effectiveness. A stricter threshold for the CNN branch is essential to avoid introducing noise pseudo labels into the ViT branch, which in turn significantly improves the ViT branch's performance. Nevertheless, opting for a high threshold for both two branches can lead to overlooking significant information in the unlabeled target domain, rendering it less effective in addressing data bias. As a result, we have chosen the thresholds of  $\tau_{vit} = 0.6$ and $\tau_{cnn} = 0.9$ for all our experiments.

\begin{table*}[!t]
\resizebox{\textwidth}{!}{%
\begin{tabular}{c|cccccccccccc|c} \toprule
Method & $rel{\rightarrow}clp$ & $rel{\rightarrow}pnt$ & $rel{\rightarrow}skt$ & $clp{\rightarrow}rel$ & $clp{\rightarrow}pnt$ & $clp{\rightarrow}skt$ & $pnt{\rightarrow}rel$ & $pnt{\rightarrow}clp$ & $pnt{\rightarrow}skt$ & $skt{\rightarrow}rel$ & $skt{\rightarrow}clp$ & $skt{\rightarrow}pnt$ & Mean \\ \midrule

MCD \cite{MCD}              & 62.0          & 69.3          & 56.3          & 79.8          & 56.6          & 53.7          & 83.4          & 58.3          & 61.0          & 81.7          & 56.3          & 66.8          & 65.4 \\

JAN \cite{JAN}              & 65.6          & 73.6          & 67.6          & 85.0          & 65.0          & 67.2          & 87.1          & 67.9          & 66.1          & 84.5          & 72.8          & 67.5          & 72.5 \\

DANN \cite{DANN}            & 63.4          & 73.6          & 72.6          & 86.5          & 65.7          & 70.6          & 86.9          & 73.2          & 70.2          & 85.7          & 75.2          & 70.0          & 74.5 \\

COAL \cite{COAL}            & 73.9          & 75.4          & 70.5          & 89.6          & 70.0          & 71.3          & 89.8          & 68.0          & 70.5          & 88.0          & 73.2          & 70.5          & 75.9 \\

InstaPBM \cite{InstaPBM}    & 80.1          & 75.9          & 70.8          & 89.7          & 70.2          & 72.8          & 89.6          & 74.4          & 72.2          & 87.0          & 79.7          & 71.8          & 77.8 \\

SENTRY \cite{SENTRY}        & 83.9          & 76.7          & 74.4          & 90.6          & 76.0           & 79.5         & 90.3          & 82.9          & 75.6          & \underline{90.4}          & 82.4          & 74.0          & 81.4 \\

RHWD \cite{RHWD}            & \textbf{84.8}          & 76.9          & 75.2          & \underline{91.8}          & 75.6          & \underline{81.2}          & \textbf{91.9}          & \textbf{84.6}          & 76.1          & \textbf{91.3}          & \underline{83.2}          & 74.6          & 82.0 \\

GSDE \cite{GSDE}            & 82.9          & \underline{79.2}          & \textbf{80.8}          & \textbf{91.9}          & \underline{78.2}          & 80.0          & 90.9          & \underline{84.1}          & \underline{79.2}          & 90.3          & \textbf{83.4}          & \underline{76.1}          & \underline{83.1} \\

\textbf{ECB(CNN)}           & \underline{84.7}          & \textbf{83.8}          & \underline{79.7}          & 91.6         & \textbf{84.0}          & \textbf{82.5}          & \underline{91.0}          & 83.2          & \textbf{79.2}          & 86.1          & 82.9          & \textbf{81.6}          & \textbf{84.2} \\ \bottomrule
\end{tabular}
}
\caption{\textbf{Accuracy (\%) on DomainNet} of UDA methods. \textbf{ECB (CNN)} represents the performance of our CNN branch when applied to ResNet-50. To facilitate easy identification, the best and second-best accuracy results are highlighted in \textbf{bold} and \underline{underline}, respectively.} \vspace{-0.25em} \label{result_domainnet_uda}
\end{table*}

\begin{table*}[!t]
\resizebox{\textwidth}{!}{%
\begin{tabular}{c|cccccccccccc|c} \toprule
Method & $A{\rightarrow}C$ & $A{\rightarrow}P$ & $A{\rightarrow}R$ & $C{\rightarrow}A$ & $C{\rightarrow}P$ & $C{\rightarrow}R$ & $P{\rightarrow}A$ & $P{\rightarrow}C$ & $P{\rightarrow}$R & $R{\rightarrow}A$ & $R{\rightarrow}C$ & $R{\rightarrow}P$ & Mean \\ \midrule
\multicolumn{14}{c}{1-shot}   \\ \midrule



ENT \cite{ENT}       &     52.9      &     75.0      &     76.7      &     63.2      &     73.6      &     73.2      &     63.0      &     51.9      &     79.9      &     70.4      &     53.6      &     81.9      &     67.9      \\

MME \cite{MME}       &     59.6      &     75.5      &     77.8      &     65.7      &     74.5      &     74.8      &     64.7      &     57.4      &     79.2      &     71.2      &     61.9      &     82.8      &     70.4      \\


DECOTA \cite{DECOTA} &     42.1      &     68.5      &     72.6      &     60.3      &     70.4      &     70.7      &     60.0      &     48.8      &     76.9      &     71.3      &     56.0      &     79.4      &     64.8      \\

CDAC \cite{CDAC}     &     61.2      &     75.9      &     78.5      &     64.5      &     75.1      &     75.3      &     64.6      &     59.3      &     80.0      &     72.7      &      61.9     &     83.1      &     71.0      \\

CDAC+SLA \cite{SLA}  &     63.0      &     78.0      &     79.2      &     66.9      &      77.6     &     77.0      &     67.3      & \underline{61.8} &     80.5      &     72.7      &     66.1      &     84.6      &     72.9      \\ 


ProML \cite{ProML}   & \underline{64.5} & \underline{79.7} & \underline{81.7} & \underline{69.1} & \underline{80.5} & \underline{79.0} & \underline{69.3} &     61.4      & \underline{81.9} & \underline{73.7} & \underline{67.5} & \underline{86.1} & \underline{74.6} \\ 

\textbf{ECB (CNN)}  & \textbf{72.9} & \textbf{88.3} & \textbf{89.6} & \textbf{84.8} & \textbf{91.3} & \textbf{89.5} & \textbf{82.9} & \textbf{71.2} & \textbf{89.9} & \textbf{85.5} & \textbf{75.4} & \textbf{92.0} & \textbf{84.4}   \\  \midrule


\multicolumn{14}{c}{3-shot}   \\ \midrule



ENT \cite{ENT}       &     61.3      &     79.5      &     79.1      &     64.7      &     79.1      &     76.4      &     63.9      &     60.5      &     79.9      &     70.2      &     62.6      &     85.7      &     71.9      \\

MME \cite{MME}       &     63.6      &     79.0      &     79.7      &     67.2      &     79.3      &     76.6      &     65.5      &     64.6      &     80.1      &     71.3      &     64.6      &     85.5      &     73.1      \\

DECOTA \cite{DECOTA} &     64.0      &     81.8      &     80.5      &     68.0      &     83.2      &     79.0      &     69.9      &     68.0      &     82.1      &     74.0      &     70.4      &     87.7      &     75.7      \\

CDAC \cite{CDAC}     &     65.9      &     80.3      &     80.6      &     67.4      &     81.4      &     80.2      &     67.5      &     67.0      &     81.9      &     72.2      &     67.8      &     85.6      &     74.8      \\



CDAC+SLA \cite{SLA}  &     67.3      &     82.6      &     81.4      &     69.2      &     82.1      &     80.1      &     70.1      & \underline{69.3} &     82.5      &     73.9      &     70.1      &     87.1      &     76.3      \\ 


ProML \cite{ProML}  & \underline{67.8} & \underline{83.9} & \underline{82.2} & \underline{72.1} & \underline{84.1} & \underline{82.3} & \underline{72.5} &     68.9      & \underline{83.8} & \underline{75.8} & \underline{71.0} & \underline{88.6} & \underline{77.8} \\

\textbf{ECB (CNN)}  & \textbf{78.7} & \textbf{90.2} & \textbf{91.3} & \textbf{85.2} & \textbf{90.4} & \textbf{91.0} & \textbf{83.9} & \textbf{76.8} & \textbf{91.2} & \textbf{85.6} & \textbf{77.6} & \textbf{92.8} & \textbf{86.2}  \\ \bottomrule

\end{tabular} 
} \vspace{-0.25em}
\caption{\textbf{Accuracy (\%) on Office-Home} of SSDA methods using a ResNet-34 serving as a backbone across different domain shifts.} \vspace{-0.95em} \label{result_officehome} 
\end{table*}

\section{Additional Experiment Results}

\subsection{Experiments Setup}

\textbf{Dataset Details.} \textit{DomainNet} is one of the largest and most diverse datasets for domain adaptation. It contains $596,010$ images distributed across $345$ categories and 6 domains: Clipart ($clp$), Infograph ($inf$), Painting ($pnt$), Quickdraw ($qdr$), Real ($rel$), and Sketch ($skt$). In the context of unsupervised domain adaptation (UDA), we encounter significant labeling noise within its full version. This is particularly evident in some classes on certain domains with many mislabeled outliers, as demonstrated in COAL \cite{COAL} and SENTRY \cite{SENTRY}. Rather than using the full set, we opt for a subset from \textit{DomainNet} featuring $40$ frequently observed classes across 4 domains: $rel$, $clp$, $pnt$,  and $skt$, encompassing all 12 possible domain shifts. In the context of SSDA, a subset of the \textit{DomainNet} dataset has been selected, focusing specifically on $126$ categories out of the original $345$. The reduced number of categories in this subset still encompasses a wide range of objects and themes, ensuring that the dataset remains complex and challenging for SSDA research. \textit{Office-Home} \cite{Office-Home} is a diverse dataset designed for domain adaptation and transfer learning research, containing around $15,500$ images from $65$ categories of everyday objects. It includes 4 significantly different domains: Art ($A$), Clipart ($C$), Product ($P$), and Real World ($R$). This variety in domains provides a challenging testbed for algorithms aiming to generalize across different visual domains. \textit{Office-31} \cite{Office-31} is an earlier standard dataset for domain adaptation,  which includes $4,110$ images across $31$ categories collected from an office environment. It consists of three distinct domains: Amazon ($A$), with $2,817$ images from amazon.com product listings; Webcam ($W$), consisting of $795$ images taken with a webcam; and DSLR ($D$), which includes $498$ images captured with a digital SLR camera. Each domain presents unique challenges regarding image quality, lighting, and backgrounds.

\noindent\textbf{Implementation Details.} In this section, we delve deeper into the specifics of our implementation. Our hybrid model utilizes the ViT/B-16 \cite{ViT} for the ViT encoder $E_{1}$. The ResNet \cite{ResNet-34} and AlexNet \cite{AlexNet} for the CNN encoder $E_{2}$. These all are initialed pre-training on the ImageNet-1K \cite{ImageNet}. Specifically, in the context of unsupervised domain adaptation (UDA), we have chosen ResNet-50 as our primary network for $E_2$, aligning with methodologies in prior studies \cite{COAL, MCD, SENTRY, DANN, RHWD}. Following the evaluation protocol of established SSDA methods \cite{ENT, MME, CDAC, G-ABC}, we employ ResNet-34 to evaluate on both the \textit{DomainNet} and \textit{Office-Home} dataset, while AlexNet is chosen for \textit{Office-31} evaluations. We are following ViT encoder $E_{1}$ and CNN encoder $E_{2}$ by two different classifiers, $F_1$ and $F_2$, each consisting of two fully-connected layers followed by the softmax function. Our ECB approach uses stochastic gradient descent as the optimizer for two branches, maintaining a momentum of $0.9$ and a weight decay of $0.0005$. Acknowledging the distinct architectures of the ViT and CNN branches, we initially set their learning rates at $1e-4$ and $1e-3$, respectively. Following \cite{MME}, we employ the learning rate scheduler with the gamma and power parameters set to $1e-4$ and $0.75$, respectively. We set the same mini-batch to $32$ for all labeled and unlabeled samples. Due to ViT's outstanding properties, ${vit}\rightarrow{cnn}$ needs to be provided with more information for the CNN branch, leading to the confidence threshold for pseudo-label selection at $\tau_{vit}=0.6$. To prevent the CNN branch from introducing noise for ViT, we set a higher threshold $\tau_{cnn}=0.9$ to get reliable pseudo labels. The warmup phase for both branches on \(\mathcal{D}_l\) undergoes a fine-tuning process across $100,000$ iterations. Subsequently, we train $50,000$ iterations for our approach.

\subsection{Comparison Results}

\textbf{Results on \textit{DomainNet} under UDA setting.} We conduct a series of $12$ experiments on the subset of \textit{DomainNet}. As detailed in \cref{result_domainnet_uda}, the ECB (CNN) outperforms the SOTA method GSDE \cite{GSDE} by an increased margin of +1.1\% in average accuracy, with an overall accuracy of 84.2\%. Additionally, our method significantly surpasses others in several specific domain transitions. Notably $rel{\rightarrow}pnt$, $clp{\rightarrow}pnt$ and $skt{\rightarrow}pnt$ improve accuracy by +4.6\%, +5.8\%, and +5.5\%, compared to the second-best. However, we face obstacles in the $skt{\rightarrow}rel$ transition, where our method's accuracy is -5.2\% lower than the RHWD \cite{RHWD} method.


\noindent\textbf{Results on \textit{Office-Home} under SSDA setting.} The performance of our method on the \textit{Office-Home} dataset, under both 1-shot and 3-shot settings, is showcased in \cref{result_officehome}. The results clearly demonstrate that our classification outcomes exceed prior methods in all domain adaptation scenarios presented. Notably, the ECB method improves an average classification accuracy that surpasses the nearest-competitor ProML \cite{ProML} by a notable +9.8\% in the 1-shot setting. Furthermore, we continue to impress with an average accuracy increase of +8.4\% under the 3-shot setting.

\noindent\textbf{Results on \textit{Office-31} under SSDA setting.} We use AlexNet as a backbone followed previous SSDA methods \cite{MME, CDAC, MVCL, G-ABC}. As demonstrated in \cref{result_office31}, our method consistently outperforms all other domain adaptation scenarios regarding classification results on the target set. Remarkably, our proposed method achieves an average classification accuracy of 84.6\% under the 3-shot setting. This result surpasses the nearest method G-ABC \cite{G-ABC} by +12.6\%. Furthermore, our method maintains a competitive edge with a +10.3\% higher performance even in the 1-shot setting. Reveals that our approach is not significantly affected by the CNN encoder architecture.


\begin{table}[!t]
\begin{adjustbox}{width=\columnwidth,center}
\begin{tabular}{c|cccc|cc}
\toprule
&
  \multicolumn{2}{c}{$W{\rightarrow}A$} &
  \multicolumn{2}{c|}{$D{\rightarrow}A$} &
  \multicolumn{2}{c}{Mean} \\ \vspace{0.1em}
  \multirow{-2}{*}{Method} &
  \multicolumn{1}{l}{1\textsubscript{shot}} &
  \multicolumn{1}{l}{3\textsubscript{shot}} &
  \multicolumn{1}{l}{1\textsubscript{shot}} &
  \multicolumn{1}{l|}{3\textsubscript{shot}} &
  \multicolumn{1}{l}{1\textsubscript{shot}} &
  \multicolumn{1}{l}{3\textsubscript{shot}} \\ \midrule

ENT \cite{ENT}                     & 50.7            & 64.0          & 50.0           & 66.2          & 50.4           & 65.1          \\
MME \cite{MME}                     & 57.2           & 67.3          & 55.8           & 67.8          & 56.5           & 67.6          \\
STar \cite{Star}                    & 59.8           & 69.1          & 56.8           & 69.0          & 58.3           & 69.1          \\
MVCL \cite{MVCL}                     & 56.7           & 69.0          & 59.3           & 69.1          & 58.0           & 69.1         \\
CDAC \cite{CDAC}                     & 63.4           & 70.1          & 62.8           & 70.0          & 63.1           & 70.0         \\
G-ABC \cite{G-ABC}                   & \underline{67.9}        & \underline{71.0}       & \underline{65.7} & \underline{73.1}   & \underline{66.8}     & \underline{72.0}          \\     

\textbf{ECB (CNN)}                   & \textbf{77.9}           & \textbf{85.2}          &    \textbf{76.3}           & \textbf{84.0}          & \textbf{77.1}           & \textbf{84.6}      \\ \bottomrule

\end{tabular}
\end{adjustbox} \vspace{-0.25em}
\caption{\textbf{Accuracy (\%) on Office-31} of SSDA methods in both 1-shot and 3-shot settings. AlexNet is used as the feature extractor for the CNN branch.} \vspace{-1.0em} \label{result_office31}
\end{table}

{
    \small
    \bibliographystyle{ieeenat_fullname}
    \bibliography{main}
}